\documentclass{article}

\usepackage[nonatbib]{mysty}
\usepackage[utf8]{inputenc} 
\usepackage[T1]{fontenc}    
\usepackage{hyperref}       
\usepackage{url}            
\usepackage{booktabs}       
\usepackage{amsfonts}       
\usepackage{nicefrac}       
\usepackage{microtype}      
\usepackage{bm}
\usepackage{amsmath}
\usepackage{mathtools}
\usepackage{tabularx,graphicx}
\usepackage{bbm}
\usepackage{pdflscape}
\usepackage{rotating}
\usepackage[bottom]{footmisc}
\usepackage[font=small]{subcaption}
\DeclareCaptionFont{mdit}{\itshape}
\usepackage[font=small,textfont=mdit]{caption}
\usepackage{tabularx}
\usepackage{float}
\usepackage{empheq}
\usepackage{multicol}
\usepackage{xfrac}    
\usepackage{nicefrac}  
\usepackage{diagbox}
\usepackage{graphicx}
\usepackage[]{algorithm2e}
\usepackage{times}
\usepackage{tikz}
\usepackage{natbib}
\usepackage{tabulary}
\usepackage{multirow}
\usepackage{enumitem}

\usepackage[font=small]{subcaption}
\DeclareCaptionFont{mdit}{\itshape}
\usepackage[font=small,textfont=mdit]{caption}
\usepackage[font=small,textfont=mdit]{subcaption}
\def\mathclap#1{\text{\hbox to 0pt{\hss$\mathsurround=0pt#1$\hss}}}
\newcommand{\myparagraph}[1]{\vspace{-1mm}\paragraph{#1}}
\newcommand{\mysection}[1]{\vspace{-5mm}\section{#1}\vspace{-2mm}}
\newcommand{\mysubsection}[1]{\vspace{-2mm}\subsection{#1}\vspace{-1mm}}
\newcommand{\mycaption}[1]{\vspace{-1mm}\caption{#1}\vspace{-1mm}}
\newcommand{\rulesep}{\unskip\ \vrule\ }
\usepackage{pdfpages}
\usetikzlibrary{bayesnet}
\title{Multi-Level Variational Autoencoder: \\
Learning Disentangled Representations from \\
Grouped Observations}
\author{
  Diane Bouchacourt \\
  OVAL Group\\
  University of Oxford\thanks{The work was performed as part of an internship at Microsoft Research.}\\
  \texttt{diane@robots.ox.ac.uk}\\
  \And
  Ryota Tomioka, Sebastian Nowozin\\
  Machine Intelligence and Perception Group\\
  Microsoft Research\\
  Cambridge, UK\\
  \texttt{\{ryoto,Sebastian.Nowozin\}@microsoft.com}\\
}

\begin{document}
\maketitle
\vspace{-5mm}
\begin{abstract}
\vspace{-5mm}
We would like to learn a representation of the data which decomposes an
observation into factors of variation which we can independently control.
Specifically, we want to use minimal supervision to learn a latent
representation that reflects the semantics behind a specific grouping of the
data, where within a group the samples share a common factor of variation. For
example, consider a collection of face images grouped by identity. We wish to
anchor the semantics of the grouping into a relevant and disentangled
representation that we can easily exploit. However, existing deep
probabilistic models often assume that the observations are independent and
identically distributed. We present the Multi-Level Variational Autoencoder
(ML-VAE), a new deep probabilistic model for learning a disentangled
representation of a set of grouped observations. The ML-VAE separates the
latent representation into semantically meaningful parts by working both at
the group level and the observation level, while retaining efficient test-time
inference. Quantitative and qualitative evaluations show that the ML-VAE
model (i) learns a semantically meaningful disentanglement of grouped data, (ii)
enables manipulation of the latent representation, and (iii) generalises to
unseen groups. 

\end{abstract}
\vspace{-3mm}
\mysection{Introduction}
\label{sec::intro}

\emph{Representation learning} refers to the task of learning a representation
of the data that can be easily exploited, see~\cite{Bengio:2013}.
In this work, our goal is to build a model that disentangles the data into
separate salient factors of variation and easily applies to a variety of tasks
and different types of observations.
Towards this goal there are multiple difficulties.
\emph{First}, the representative power of the learned representation depends
on the information one wishes to extract from the data.
\emph{Second}, the multiple factors of variation impact the observations in a
complex and correlated manner.
\emph{Finally}, we have access to very little, if any, supervision over these
different factors.
If there is no specific meaning to embed in the desired representation, the
\emph{infomax principle}, described in~\cite{Linsker1998}, states that an
optimal representation is one of bounded entropy which retains as much
information about the data as possible.
However, we are interested in learning a semantically meaningful
disentanglement of interesting latent factors.
How can we anchor semantics in high-dimensional representations?

We propose \emph{group-level supervision}: observations are organised in
groups, where within a group the observations share a common but unknown value
for one of the factors of variation.
For example, take images of circle and stars, of possible colors green, yellow
and blue. A possible grouping organises the images by shape (circled or
starred).
Group observations allow us to anchor the semantics of the data (shape and
color) into the learned representation.
Group observations are a form of weak supervision that is inexpensive to
collect.  In the above shape example, we do not need to know the factor of
variation that defines the grouping.

\begin{figure}[t]
\begin{subfigure}[t]{0.32\textwidth}
\centering
\captionsetup{width=.95\textwidth}
  \includegraphics[width=\textwidth]{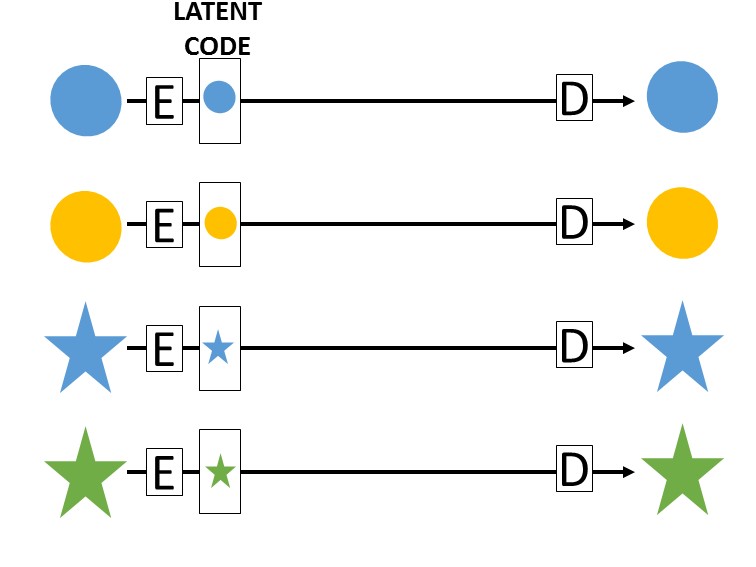}
    \vspace{-5mm}
  \mycaption{Original VAE assumes i.i.d. observations.}
  \label{fig:vaeshape}
\end{subfigure}
\rulesep
\begin{subfigure}[t]{0.32\textwidth}
\centering
\captionsetup{width=.95\textwidth}
  \includegraphics[width=\textwidth,trim={0cm 0cm 0 0},clip]{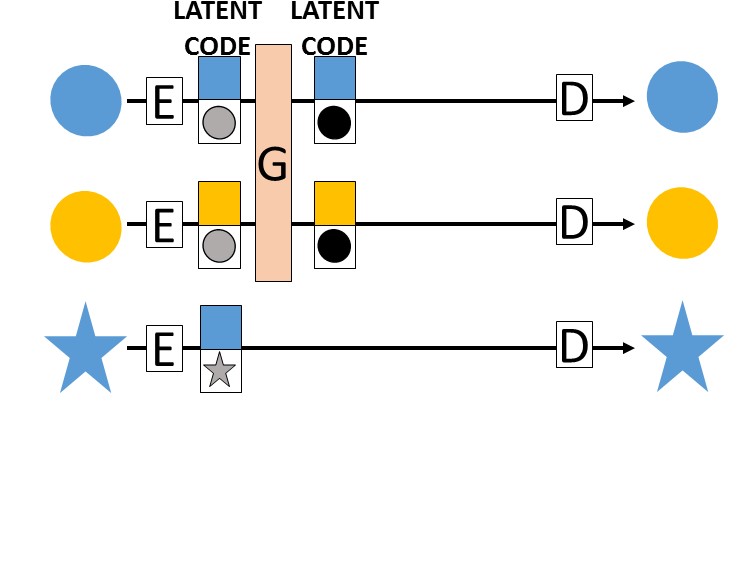}
    \vspace{-5mm}
  \mycaption{ML-VAE accumulates evidence.}
   \label{fig:multi1}
\end{subfigure}
\begin{subfigure}[t]{0.32\textwidth}
\centering
\captionsetup{width=.95\textwidth}
  \includegraphics[width=\textwidth,trim={0cm 0cm 0 0},clip]{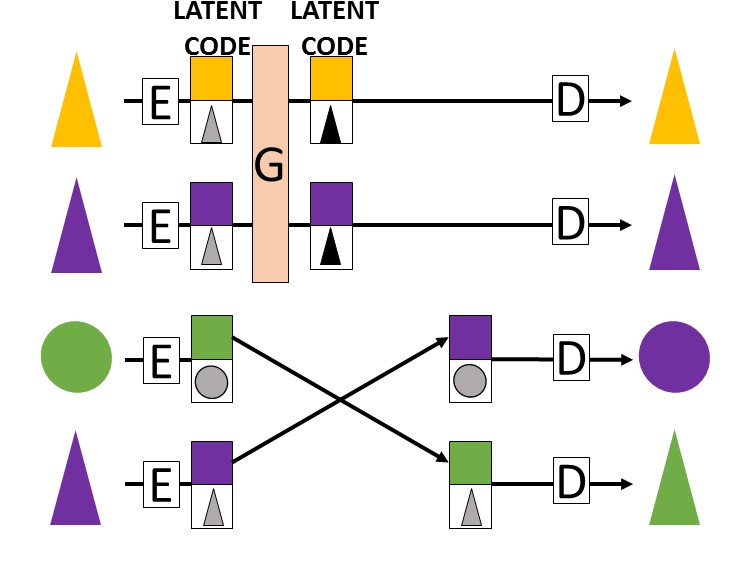}
  \vspace{-5mm}
  \mycaption{ML-VAE generalises to unseen shapes and colors and allows control on the latent code.}
   \label{fig:multi2}
\end{subfigure}
\mycaption{In (a) the VAE of~\cite{Kingma2013, Rezende2014}, it assumes i.i.d. observations. In comparison, (b) and (c) show our ML-VAE working at the group level. In (b) and (c) upper part of the latent code is color, lower part is shape. Black shapes show the ML-VAE accumulating evidence on the shape from the two grey shapes. E is the Encoder, D is the Decoder, G is the grouping operation. Best viewed in color.}
\vspace{-5mm}
\end{figure}

Deep probabilistic generative models learn expressive representations of a
given set of observations. Among them,~\cite{Kingma2013, Rezende2014} proposed
the very successful Variational Autoencoder (VAE). In the VAE model, a network
(the encoder) encodes an observation into its latent representation (or latent
code) and a generative network (the decoder) decodes an observation from a
latent code. The VAE model performs amortised inference, that is, the
observations parametrise the posterior distribution of the latent code, and
all observations share a single set of parameters to learn. This allows
efficient test-time inference. However, the VAE model assumes that the
observations are independent and identically distributed (i.i.d.). In the case
of grouped observations, this assumption is no longer true. Considering the
toy example of objects grouped by shape, the VAE model considers and processes
each observation independently. This is shown in Figure~\ref{fig:vaeshape}.
The VAE model takes no advantage of the knowledge of the grouping.

How can we build a probabilistic model that easily incorporates this grouping
information and learns the corresponding relevant representation?
We could enforce equal representations within groups in a graphical model,
using stochastic variational inference (SVI) for approximate posterior
inference,~\cite{Hoffman2013}.
However, such model paired with SVI cannot take advantage of efficient amortised
inference. As a result, SVI requires more passes over the training data and
expensive test-time inference.
Our proposed model retains the advantages of amortised inference while using
the grouping information in a simple yet flexible manner.
~\\
~\\
We present the Multi-Level Variational Autoencoder (ML-VAE), a new deep
probabilistic model that learns a disentangled representation of a set of
grouped observations.
The ML-VAE separates the latent representation into semantically meaningful
parts by working both at the group level and the observation level.
Without loss of generality we assume that there are two latent factors,~\emph{style}
and~\emph{content}. The content is common for a group, while the style can
differ within the group. We emphasise that our approach is general in that
there can be more than two factors.
Moreover, for the same set of observations, multiple groupings are possible
along different factors of variation.
To use group observations the ML-VAE uses a grouping operation that separates
the latent representation into two parts, style and content, and samples in
the same group have the same content.
This in turns makes the encoder learn a semantically meaningful
disentanglement. This process is shown in Figure~\ref{fig:multi1}. For
illustrative purposes, the upper part of the latent code represents the style
(color) and the lower part the content (shape: circle or star).
In Figure~\ref{fig:multi1}, after being encoded the two circles share the same
shape in the lower part of the latent code (corresponding to content). The
variations within the group (style), in this case color, gets naturally
encoded in the upper part. Moreover, while the ML-VAE handles the case of a
single sample in a group, if there are multiples samples in a group the
grouping operation increases the certainty on the content. This is shown in
Figure~\ref{fig:multi1} where black circles show that the model has
accumulated evidence of the content (circle) from the two disentangled codes
(grey circles). The grouping operation does not need to know that the data are
grouped by shape nor what shape and color represent; the only supervision is
the organisation of the data in groups. At test-time, the ML-VAE generalises
to unseen realisations of the factors of variation, for example the purple
triangle in Figure~\ref{fig:multi2}. Using the disentangled representation, we
can control the latent code and can perform operations such as swapping part
of the latent representation to generate new observations, as shown in
Figure~\ref{fig:multi2}. To sum-up, our contributions are as follows.\\
\vspace{-5mm}
\begin{itemize}[noitemsep,nolistsep]
\item We propose the ML-VAE model to learn disentangled representations from
group level supervision;
\item we extend amortized inference to the case of non-iid observations;
\item we demonstrate experimentally that the ML-VAE model learns a
semantically meaningful disentanglement of grouped data;
\item we demonstrate manipulation of the latent representation and generalises
to unseen groups. 
\end{itemize}
\mysection{Related Work}
\label{sec::relwork}
\vspace{-2mm}
Research has actively focused on the development of deep probabilistic models that learn to represent the distribution of the data. Such models parametrise the learned representation by a neural network. We distinguish between two types of deep probabilistic models. Implicit probabilistic models stochastically map an input random noise to a sample of the modelled distribution. Examples of implicit models include Generative Adversarial Networks (GANs) developed by~\cite{Goodfellow2014} and kernel based models, see~\cite{Li2015, Dziugaite2015, Bouchacourt2016}. The second type of model employs an explicit model distribution and builds on variational inference to learn its parameters. This is the case of the Variational Autoencoder (VAE) proposed by~\cite{Kingma2013, Rezende2014}. Both types of model have been extended to the representation learning framework, where the goal is to learn a representation that can be effectively employed. In the unsupervised setting, the InfoGAN model of~\cite{Chen2016} adapts GANs to the learning of an interpretable representation with the use of mutual information theory, and~\cite{Wang_SSGAN2016} use two sequentially connected GANs. The~$\beta$-VAE model of~\cite{Higgins2017} encourages the VAE model to optimally use its capacity by increasing the Kullback-Leibler term in the VAE objective. This favors the learning of a meaningful representation.~\cite{Abbasnejad2016} uses an infinite mixture as variational approximation to improve performance on semi-supervised tasks. Contrary to our setting, these unsupervised models do not anchor a specific meaning into the disentanglement. In the semi-supervised setting, i.e. when an output label is partly available,~\cite{Siddharth2017} learn a disentangled representation by introducing an auxiliary variable. While related to our work, this model defines a semi-supervised factor of variation. In the example of multi-class classification, it would not generalise to unseen classes. We define our model in the grouping supervision setting, therefore we can handle unseen classes at testing.\\
\\
The VAE model has been extended to the learning of representations that are invariant to a certain source of variation. In this context~\cite{Alemi2017} build a meaningful representation by using the Information Bottleneck (IB) principle, presented by~\cite{Tishby1999}. The Variational Fair Autoencoder presented by~\cite{Louizos2016} encourages independence between the latent representation and a sensitive factor with the use of a Maximum Mean Discrepancy (MMD) based regulariser, while~\cite{Edwards2016} uses adversarial training. Finally, \cite{Chen2017} control which part of the data gets encoded by the encoder and employ an autoregressive architecture to model the part that is not encoded. While related to our work, these models require supervision on the source of variation to be invariant to. In the specific case of learning interpretable representation of images,~\cite{Kulkarni2015} train an autoencoder with minibatch where only one latent factor changes. Finally,~\cite{Mathieu2016} learn a representation invariant to a certain source of data by combining autoencoders trained in an adversarial manner.\\
\\
Multiple works perform image-to-image translation between two unpaired images collections using GAN-based architectures, see~\cite{Zhu2017, Kim2017, Yi2017, Fu2017, Taigman2016,Shrivastava2016, Bousmalis2016}, while~\cite{Liu2017} employ a combination of VAE and GANs. Interestingly, all these models require a form of weak supervision that is similar to our setting. We can think of the two unpaired images collections as two groups of observed data, sharing image type (painting versus photograph for example). Our work differs from theirs as we generalise to any type of data and number of groups. It is unclear how to extend the cited models to the setting of more than two groups and other types of data. Also, we do not employ multiple GANs models but a single VAE-type model. While not directly related to our work,~\cite{Murali2017} perform computer program synthesis using grouped user-supplied example programs, and~\cite{Allamanis2017} learn continuous semantic representations of mathematical and logical expressions. Finally we mention the concurrent recent work of~\cite{Donahue2017} which disentangles the latent space of GANs. 
\mysection{Model}
\label{sec::model}
\mysubsection{Amortised Inference with the Variational Autoencoder (VAE) Model}
We define~$\bold{X}=(X_1,...,X_N)$. In the probabilistic model framework, we assume that the observations~$\bold{X}$ are generated by~$\bold{Z}$, the unobserved (latent) variables. The goal is to infer the values of the latent variable that generated the observations, that is, to calculate the posterior distribution over the latent variables~$p(\bold{Z}|\bold{X};\theta)$, which is often intractable. The original VAE model proposed by~\cite{Kingma2013, Rezende2014} approximate the intractable posterior with the use of a variational approximation~$q(\bold{Z}|\bold{X};\phi)$, where~$\phi$ are the variational parameters. Contrary to Stochastic Variational Inference (SVI), the VAE model performs amortised variational inference, that is, the observations parametrise the posterior distribution of the latent code, and all observations share a single set of parameters~$\phi$. This allows efficient test-time inference. Figure~\ref{graphiid} shows the SVI and VAE graphical models, we highlight in red that the SVI model does not take advantage of amortised inference.
\begin{figure}[t]
\begin{subfigure}[b]{0.5\textwidth}
\centering
\begin{tikzpicture}
\tikzstyle{main}=[circle, minimum size = 7mm, thick, draw =black!80, node distance = 7mm]
\tikzstyle{connect}=[-latex]
\tikzstyle{box}=[rectangle, draw=black!100]
  \node[main,fill=gray] (x_i) [label=center:$X_i$] {};
  \node[main] (z) [above=of x_i,label=center:$Z_{i}$] {};
  \node[main,draw=none,node distance = 5mm] (phi) [left=of z, label=center:\color{red}\large $\phi_i$] {};
  \node[main,draw=none,node distance = 20mm] (theta) [left=of x_i, label=center:\large $\theta$] {};
  \node[draw=none,node distance = 4mm] (N) [below=of x_i, label=center:$i \in {[1,N]}$] {};
  \path (phi) edge [connect, dashed, draw=black!100, line width=0.1mm] (z);
  \path (z) edge[connect, draw=black!100, line width=0.1mm] (x_i);
  \path (theta) edge[connect, draw=black!100, line width=0.1mm]  (x_i);
  \node[rectangle, inner sep=4mm, draw=black!100, fit= (phi)(x_i)(z)(N)] {};
\end{tikzpicture}
\mycaption{SVI graphical model.}
\label{graph::svi}
\end{subfigure}
\begin{subfigure}[b]{0.5\textwidth}
\centering
\begin{tikzpicture}
\tikzstyle{main}=[circle, minimum size = 7mm, thick, draw =black!80, node distance = 7mm]
\tikzstyle{connect}=[-latex]
\tikzstyle{box}=[rectangle, draw=black!100]
  \node[main,fill=gray] (x_i) [label=center:$X_i$] {};
  \node[main] (z) [above=of x_i,label=center:$Z_{i}$] {};
  \node[main,draw=none,node distance = 8mm] (phi) [left=of z, label=center:\large $\phi$] {};
  \node[main,draw=none,node distance = 8mm] (theta) [left=of x_i, label=center:\large $\theta$] {};
  \node[draw=none,node distance = 4mm] (N) [below=of x_i, label=center:$i \in {[1,N]}$] {};
  \path (x_i) edge [connect, dashed, bend left, draw=black!100, line width=0.1mm] (z);
  \path (phi) edge [connect, dashed, draw=black!100, line width=0.1mm] (z);
  \path (z) edge[connect, draw=black!100, line width=0.1mm]  (x_i);
  \path (theta) edge[connect, draw=black!100, line width=0.1mm]  (x_i);
  \node[rectangle, inner sep=4mm, draw=black!100, fit= (x_i)(z)(N)] {};
\end{tikzpicture}
\mycaption{VAE graphical model.}
\label{graph::vae}
\end{subfigure}
\mycaption{VAE~\cite{Kingma2013, Rezende2014} and SVI~\cite{Hoffman2013} graphical models. Solid lines denote the generative model, dashed lines denote the variational approximation.}
\label{graphiid}
\vspace{-5mm}
\end{figure}

\mysubsection{The ML-VAE for Grouped Observations}
We now assume that the observations are organised in a set~$\mathcal{G}$ of distinct groups, with a factor of variation that is shared among all observations within a group. The grouping forms a partition of~$[1,N]$, i.e. each group~$G \in \mathcal{G}$ is a subset of~$[1,N]$ of arbitary size, disjoint of all other groups. Without loss of generality, we separate the latent representation in two latent variables~$Z=(C,S)$ with~\emph{style}~$S$ and~\emph{content}~$C$. The content is the factor of variation along which the groups are formed. In this context, referred as the grouped observations case, the latent representation has a single content latent variable per group~$C_G$. SVI can easily be adapted by enforcing that all observations within a group share a single content latent variable while the style remains untied, see Figure~\ref{graph::svigroup}. However, employing SVI requires iterative test-time inference since it does not perform amortised inference. Experimentally, it also requires more passes on the training data as we show in the supplementary material. The VAE model assumes that the observations are i.i.d, therefore it does not take advantadge of the grouping. In this context, the question is how to perform amortised inference in the context of non-i.i.d., grouped observations? In order to tackle the aforementioned deficiency we propose the Multi-Level VAE (ML-VAE).\\
\\
We denote by~$\bold{X}_G$ the observations corresponding to the group~$G$. We explicitly model each~$X_i$ in~$\bold{X}_G$ to have its independent latent representation for the style~$S_i$, and~$\bold{S}_{G}=(S_i, i \in G)$. $C_G$ is a unique latent variable shared among the group for the content. The variational approximation~$q(C_G,\bold{S}_G|\bold{X}_G;\phi)$ factorises and $\phi_c$ and~$\phi_s$ are the variational parameters for content and style respectively. We assume that the style is independent in a group, so~$\bold{S}_{G}$ also factorises. Finally, given style and content, the likelihood~$p(\bold{X}_G|C_G,\bold{S}_G;\theta)$ decomposes on the samples. This results in the graphical model shown Figure~\ref{graph::ours}.\\
\begin{figure}[h!]
\begin{subfigure}[b]{0.5\textwidth}
\centering
\begin{tikzpicture}
\tikzstyle{main}=[circle, minimum size = 7mm, thick, draw =black!80, node distance = 7mm]
\tikzstyle{connect}=[-latex]
\tikzstyle{box}=[rectangle, draw=black!100]
  \node[main,fill=gray] (x_i) [label=center:$X_i$] {};
  \node[main] (z_s) [above=of x_i,label=center:$S_i$] {};
  \node[main, node distance = 5mm] (z_c) [right=of z_s,label=center:$C_G$] {};
  \node[main, draw=none,node distance = 5mm] (phi_s) [left=of z_s, label=center:\color{red}\large $\phi_{s,i}$] {};
  \node[main,draw=none,node distance = 5mm] (phi_c) [right=of z_c, label=center:\color{red}\large $\phi_{c,G}$] {};
  \node[main,draw=none,node distance = 20mm] (theta) [left=of x_i, label=center:\large $\theta$] {};
  \node[draw=none, node distance = 4mm] (G) [below=of x_i, label=center:$i \in G$] {};
  \path (phi_s) edge [connect, dashed, draw=black!100, line width=0.1mm]  (z_s);
  \path (phi_c) edge [connect, dashed, draw=black!100, line width=0.1mm] (z_c);
  \path (z_c) edge[connect, draw=black!100, line width=0.1mm]  (x_i);
  \path (z_s) edge[connect, draw=black!100, line width=0.1mm]  (x_i);
  \path (theta) edge[connect,draw=black!100, line width=0.1mm]  (x_i);
  \node[rectangle, inner sep=3mm, draw=black!100, fit= (phi_s)(x_i)(z_s)(G)] {};
  \node[rectangle, inner sep=5mm, draw=black!100, fit= (phi_s)(phi_c)(x_i)(z_s)(z_c)(G)] {};
  \node[rectangle, inner sep=5mm, fit= (x_i)(z_s)(z_c)(G),label=below right:$G \in \mathcal{G}$,xshift=0mm] {};
\end{tikzpicture}
\mycaption{SVI for grouped observations.}
\label{graph::svigroup}
\end{subfigure}
\begin{subfigure}[b]{0.50\textwidth}
\centering
\begin{tikzpicture}
\tikzstyle{main}=[circle, minimum size = 7mm, thick, draw =black!80, node distance = 7mm]
\tikzstyle{connect}=[-latex, thick]
\tikzstyle{box}=[rectangle, draw=black!100]
  \node[main,fill=gray] (x_i) [label=center:$X_i$] {};
  \node[main] (z_s) [above=of x_i,label=center:$S_i$] {};
  \node[main, node distance = 5mm] (z_c) [right=of z_s,label=center:$C_G$] {};
  \node[main,draw=none,node distance = 8mm] (phi_s) [left=of z_s, label=center:\large $\phi_{s}$] {};
  \node[main,draw=none,node distance = 8mm] (phi_c) [right=of z_c, label=center:\large $\phi_{c}$] {};
  \node[main,draw=none,node distance = 8mm] (theta) [left=of x_i, label=center:\large $\theta$] {};
  \node[draw=none, node distance = 4mm] (G) [below=of x_i, label=center:$i \in G$] {};
  \path (phi_s) edge [connect, dashed, draw=black!100,line width=0.1mm] (z_s);
  \path (phi_c) edge [connect, dashed, draw=black!100,line width=0.1mm] (z_c);
  \path (x_i) edge [connect, dashed, bend left=20, draw=black!100,line width=0.1mm] (z_s);
  \path (x_i) edge [connect, dashed, bend right=20, draw=black!100,line width=0.1mm] (z_c);
  \path (z_c) edge[connect, draw=black!100,line width=0.1mm]  (x_i);
  \path (z_s) edge[connect, draw=black!100,line width=0.1mm]  (x_i);
  \path (theta) edge[connect,draw=black!100,line width=0.1mm]  (x_i);
  \node[rectangle, inner sep=3mm, draw=black!100, fit= (x_i)(z_s)(G)] {};
  \node[rectangle, inner sep=5mm, draw=black!100, fit= (x_i)(z_s)(z_c)(G)] {};
  \node[rectangle, inner sep=5mm, fit= (x_i)(z_s)(z_c)(G),label=below right:$G \in \mathcal{G}$,xshift=-12mm] {};
\end{tikzpicture}
\mycaption{Our ML-VAE.}
\label{graph::ours}
\end{subfigure}
\label{graph::groupcase}
\mycaption{SVI~\cite{Hoffman2013} and our ML-VAE graphical models. Solid lines denote the generative model, dashed lines denote the variational approximation.}
\end{figure}
We do not assume i.i.d. observations, but independence at the grouped observations level. The average marginal log-likelihood decomposes over groups of observations
\begin{equation}
\begin{split}
\dfrac{1}{|\mathcal{G}|}\log p(\bold{X}|\theta)=\dfrac{1}{|\mathcal{G}|} \sum_{G \in \mathcal{G}} \log p(\bold{X}_G|\theta).
\end{split}
\label{trueobj}
\end{equation}
For each group, we can rewrite the marginal log-likelihood as the sum of the group Evidence Lower Bound~$\textrm{ELBO}(G; \theta, \phi_s, \phi_c)$ and the Kullback-Leibler divergence between the true posterior~$p(C_G,\bold{S}_G|\bold{X}_G;\theta)$ and the variational approximation~$q(C_G,\bold{S}_G|\bold{X}_G;\phi_c)$. Since this Kullback-Leibler divergence is always positive, the first term,~$\textrm{ELBO}(G; \theta, \phi_s, \phi_c)$, is a lower bound on the marginal log-likelihood,
\begin{equation}
\begin{split}
\log p(\bold{X}_G|\theta)&=\textrm{ELBO}(G; \theta, \phi_s, \phi_c) + \textrm{KL}(q(C_G,\bold{S}_G|\bold{X}_G;\phi_c)||p(C_G,\bold{S}_G|\bold{X}_G;\theta))\\
&\ge~\textrm{ELBO}(G; \theta, \phi_s, \phi_c).
\end{split}
\end{equation}
The~$\textrm{ELBO}(G; \theta, \phi_s, \phi_c)$ for a group is
\begin{equation}
\begin{split}
\textrm{ELBO}(G; \theta, \phi_s, \phi_c) &=  \sum_{i \in G} \mathbb{E}_{q(C_G|\bold{X}_G;\phi_c)}[\mathbb{E}_{q(S_i|X_i;\phi_s)}
[\log p(X_i|C_G,S_i;\theta)]]\\
&\quad - \sum_{i \in G}\textrm{KL}(q(S_i|X_i;\phi_s)||p(S_i)) -\textrm{KL}(q(C_G|\bold{X}_G;\phi_c)||p(C_G)).
\end{split}
\end{equation}
We define the average group ELBO over the dataset,~$\displaystyle \mathcal{L}(\mathcal{G},\theta, \phi_c, \phi_s):=\dfrac{1}{|\mathcal{G}|}\sum_{G \in \mathcal{G}}\textrm{ELBO}(G; \theta, \phi_s, \phi_c)$
and we maximise~$\mathcal{L}(\mathcal{G}, \phi_c, \phi_s, \theta)$. It is a lower bound on~$\dfrac{1}{|\mathcal{G}|}\log p(\bold{X}|\theta)$ because each group Evidence Lower Bound~$\textrm{ELBO}(G; \theta, \phi_s, \phi_c)$ is a lower bound on~$p(\bold{X}_G|\theta)$, therefore,
\begin{equation}
\begin{split}
\dfrac{1}{|\mathcal{G}|}\log p(\bold{X}|\theta)=\dfrac{1}{|\mathcal{G}|} \sum_{G \in \mathcal{G}} \log p(\bold{X}_G|\theta)\ge \mathcal{L}(\mathcal{G}, \phi_c, \phi_s, \theta).
\end{split}
\end{equation} 
In comparison, the original VAE model maximises the average ELBO over individual samples. In practise, we build an estimate of~$\mathcal{L}(\mathcal{G},\theta, \phi_c, \phi_s)$ using minibatches of group.
\begin{equation}
\begin{split}
\mathcal{L}(\mathcal{G}_b,\theta, \phi_c, \phi_s)=\dfrac{1}{|\mathcal{G}_b|}\sum_{G \in \mathcal{G}_b}\textrm{ELBO}(G; \theta, \phi_s, \phi_c).
\end{split}
\label{minibatchobj}
\end{equation}
If we take each group~$G \in \mathcal{G}_b$, in its entirety this is an unbiased estimate. When the groups sizes are too large, for efficiency, we subsample $G$ and this estimate is biased. We discuss the bias in the supplementary material. The resulting algorithm is shown in Algorithm~\ref{algo:algtraining}.\\
\RestyleAlgo{ruled}
\SetAlgoNoLine
\LinesNumbered
\SetAlgoLined
\begin{algorithm}[b]
\For{Each epoch}{
Sample minibatch of groups~$\mathcal{G}_b$,\\
\For{$G \in \mathcal{G}_b$}{
  \For{$i \in G$}{
    Encode~$x_i$ into~$q(C_G|X_i=x_i;\phi_c)$, ~$q(S_i|X_i=x_i;\phi_s),$
  }
  Construct~$\displaystyle q(C_G|\bold{X}_G=\bold{x}_G;\phi_c)$\label{eqConstruct} using~$q(C_G|X_i=x_i;\phi_c), \forall i \in G$,\\
  \For{$i \in G$}{
    Sample~$c_{G,i} \sim q(C_G|\bold{X}_G=\bold{x}_G;\phi_c)$, $s_i \sim q(S_i|X_i=x_i;\phi_s)$ \label{sampdecode},\\
    Decode~$c_{G,i}, s_i$ to obtain~$p(X_i|C_G=c_{G,i}, S_i=s_i;\theta)$,\\
  }
}
Update~$\theta,\phi_c,\phi_s$ by taking a gradient step of Equation~\eqref{minibatchobj}: $\nabla_{\theta,\phi_c,\phi_s} \mathcal{L}(\mathcal{G}_b,\theta, \phi_c, \phi_s)$
}
\mycaption{ML-VAE training algorithm.}
\label{algo:algtraining}
\end{algorithm}
\\
For each group~$G$, in step~\ref{eqConstruct} of Algorithm~\ref{algo:algtraining} we build the group content distribution by accumulating information from the result of encoding each sample in~$G$. The question is how can we accumulate the information in a relevant manner to compute the group content distribution?

\mysubsection{Accumulating Group Evidence using a Product of Normal densities}
\label{subsec::prod}
Our idea is to build the variational approximation of the single group content variable,~$q(C_G|\bold{X}_G;\phi_c)$, from the encoding of the grouped observations~$\bold{X}_G$. While any distribution could be employed, we focus on using a product of Normal density functions. Other possibilities, such as a mixture of density functions, are discussed in the supplementary material.\\
\\
We construct the probability density function of the latent variable~$C_G$ taking the value~$c$ by multiplying~$|G|$ normal density functions, each of them evaluating the probability of~$C_G=c$ given~$X_i=x_i, i \in G$,
\begin{equation}
\begin{split}
q(C_G=c|\bold{X}_G=\bold{x}_G;\phi_c) & \propto \prod_{i \in G}q(C_G=c|X_i=x_i;\phi_c),
\end{split}
\end{equation}
where we assume~$q(C_G|X_i=x_i;\phi_c)$ to be a Normal distribution~$N(\mu_i,\Sigma_i)$. ~\cite{Murphy2007} shows that the product of two Gaussians is a Gaussian. Similarly, in the supplementary material we show that~$q(C_G=c|\bold{X}_G=\bold{x}_G;\phi_c)$ is the density function of a Normal distribution of mean~$\mu_G$ and variance~$\Sigma_G$
\begin{equation}
\begin{split}
\Sigma_G^{-1} = \sum_{i \in G}\Sigma_i^{-1},~\mu_G^T\Sigma_G^{-1} =  \sum_{i \in G}\mu_i^T\Sigma_i^{-1}.
\end{split}
\label{prodparameters}
\end{equation}
It is interesting to note that the variance of the resulting Normal distribution,~$\Sigma_G$, is inversely proportional to the sum of the group's observations inverse variances~$\sum_{i \in G}\Sigma_i^{-1}$. Therefore, we expect that by increasing the number of observations in a group, the variance of the resulting distribution decreases. This is what we refer as ``accumulating evidence''. We empirically investigate this effect in Section~\ref{sec::expe}. Since the resulting distribution is a Normal distribution, the term~$\textrm{KL}(q(C_G|\bold{X}_G;\phi_c)||p(C_G))$ can be evaluated in closed-form. We also assume a Normal distribution for~$q(S_i|X_i;\phi_s), i \in G$. 

\mysection{Experiments}
\label{sec::expe}
We evaluate the ML-VAE on images, other forms of data are possible and we leave these for future work. In all experiments we use the Product of Normal method presented in Section~\ref{subsec::prod} to construct the content latent representation. Our goal with the experiments is twofold. First, we want to evaluate the performance of ML-VAE to learn a semantically meaningful disentangled representation. Second, we want to explore the impact of ``accumulating evidence'' described in Section~\ref{subsec::prod}. Indeed when we encode test images two strategies are possible: strategy~~$1$ is disregarding the grouping information of the test samples, i.e. each test image is a group; and strategy~$2$ is considering the grouping information of the test samples, i.e. taking multiple test images per identity to construct the content latent representation. 

\myparagraph{MNIST dataset.} We evaluate the ML-VAE on MNIST~\cite{Lecun1998}. We consider the data grouped by digit label, i.e. the content latent code~$C$ should encode the digit label. We randomly separate the~$60,000$ training examples into~$50,000$ training samples and~$10,000$ validation samples, and use the standard MNIST testing set. For both the encoder and decoder, we use a simple architecture of~$2$ linear layers (detailed in the supplementary material).

\myparagraph{MS-Celeb-1M dataset.} Next, we evaluate the ML-VAE on the face aligned version of the MS-Celeb-1M dataset~\cite{Guo2016}. The dataset was constructed by retrieving approximately~$100$ images per celebrity from popular search engines, and noise has not been removed from the dataset. For each query, we consider the top ten results (note there was multiple queries per celebrity, therefore some identities have more than~$10$ images). This creates a dataset of~$98,880$ entities for a total of~$811,792$ images, and we group the data by identity. Importantly, we randomly separate the dataset in disjoints sets of identities as the training, validation and testing datasets. This way we evaluate the ability of ML-VAE level to generalise to unseen groups (unseen identities) at test-time. The training dataset consists of~$48,880$ identities (total~$401,406$ images), the validation dataset consists of~$25,000$ identities (total~$205,015$ images) and the testing dataset consists of~$25,000$ identities (total~$205,371$ images). The encoder and the decoder network architectures, composed of either convolutional or deconvolutional and linear layers, are detailed in the supplementary material. We resize the images to~$64 \times 64$ pixels to fit the network architecture. 

\myparagraph{Qualitative Evaluation.}
As explained in~\cite{Mathieu2016}, there is no standard benchmark dataset or metric to evaluate a model on its disentanglement performance. Therefore similarly to~\cite{Mathieu2016} we perform qualitative and quantitative evaluations. We qualitatively assess the relevance of the learned representation by performing operations on the latent space. First we perform swapping: we encode test images, draw a sample per image from its style and content latent representations, and swap the style between images. Second we perform interpolation: we encode a pair of test images, draw one sample from each image style and content latent codes, and linearly interpolate between the style and content samples. We present the results of swapping and interpolation with accumulating evidence of~$10$ other images in the group (strategy~$2$). Results without accumulated evidence (strategy~$1$) are also convincing and available in the supplementary material. We also perform generation: for a given test identity, we build the content latent code by accumulating images of this identity. Then take the mean of the resulting content distribution and generate images with styles sampled from the prior. Finally in order to explore the benefits of taking into account the grouping information, for a given test identity, we reconstruct all images for this identity using both these strategies and show the resulting images.
\begin{figure}[t]
\begin{subfigure}[t]{0.5\textwidth}
\centering
\captionsetup{width=.95\textwidth}
  \includegraphics[width=0.9\textwidth,trim={4cm 4cm 5cm 4cm},clip]{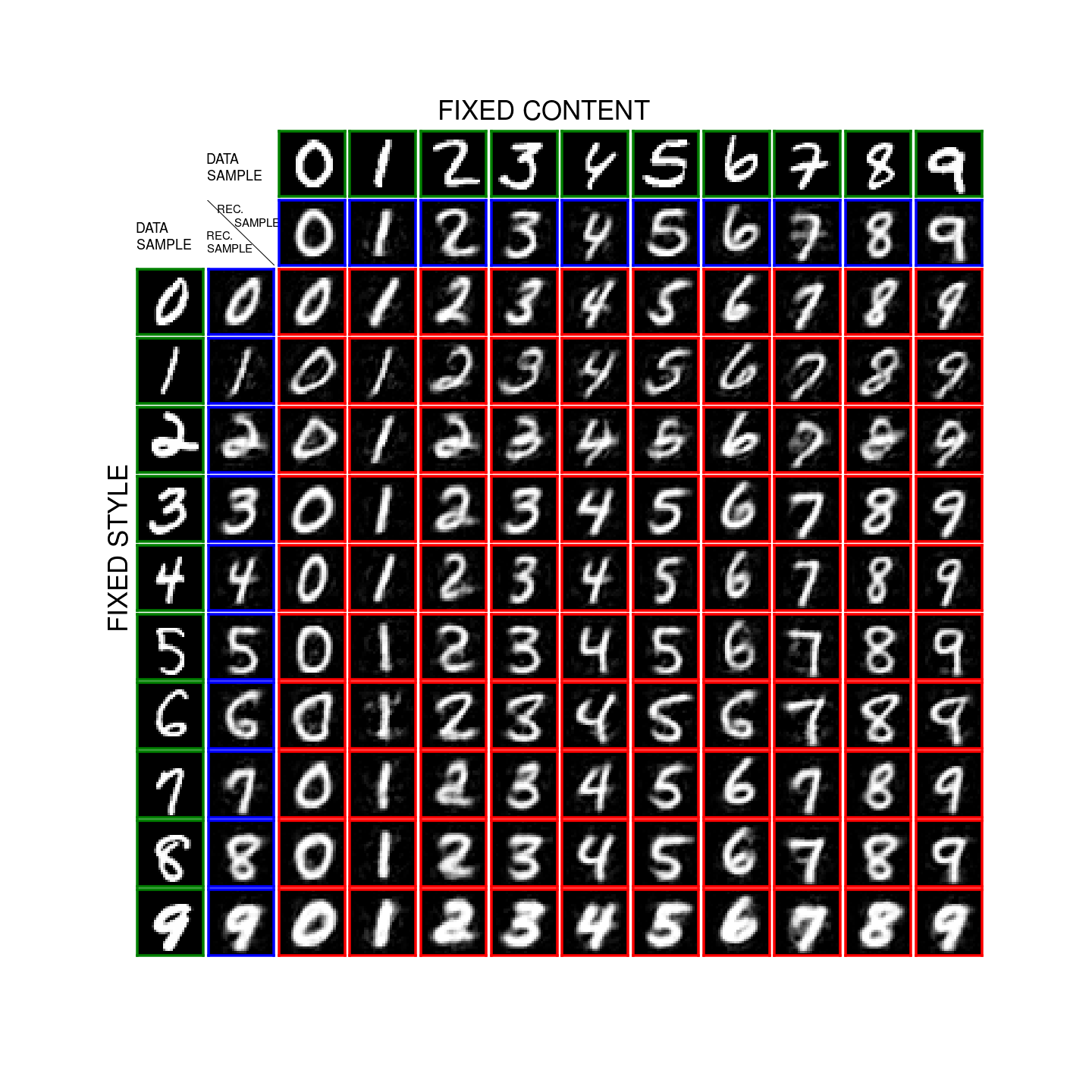}
  \vspace{-4mm}
  \mycaption{MNIST, test dataset.}
\end{subfigure}
\hfill
\begin{subfigure}[t]{0.5\textwidth}
\centering
\captionsetup{width=.95\textwidth}
  \includegraphics[width=0.9\textwidth,trim={4cm 4cm 5cm 4cm},clip]{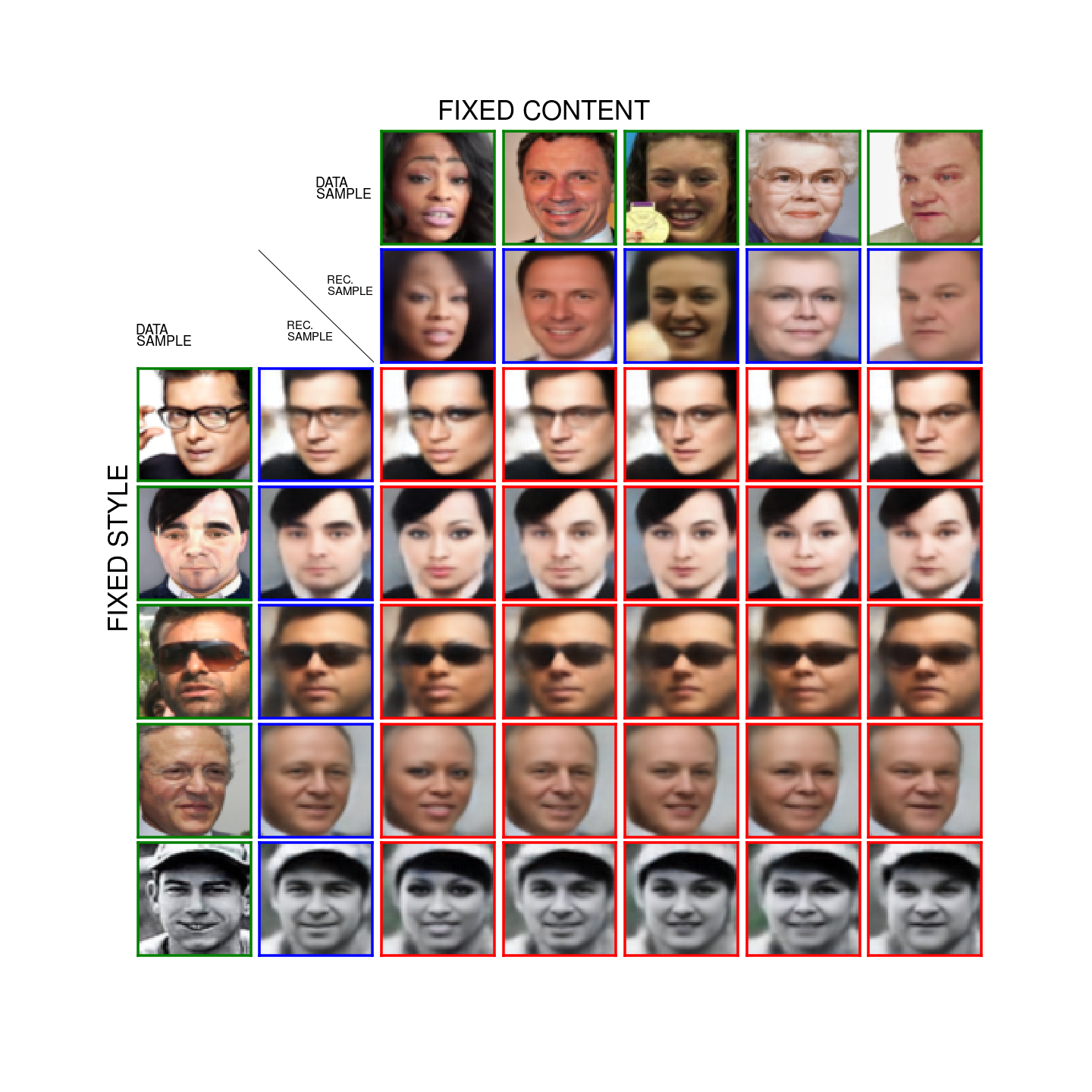}
  \vspace{-4mm}
    \mycaption{MS-Celeb-1M, test dataset.}
\end{subfigure}
\vspace{-5mm}
\mycaption{Swapping, first row and first column are test data samples (green boxes), second row and column are reconstructed samples (blue boxes) and the rest are swapped reconstructed samples (red boxes). Each row is fixed style and each column is a fixed content. Best viewed in color on screen.}
\label{swapping}
\vspace{-6mm}  
\end{figure}
\begin{figure}[b]
\begin{subfigure}[t]{0.49\textwidth}
\captionsetup{width=.95\textwidth}
\centering
  \includegraphics[width=0.8\textwidth,trim={0cm 0cm 0cm 0cm},clip]{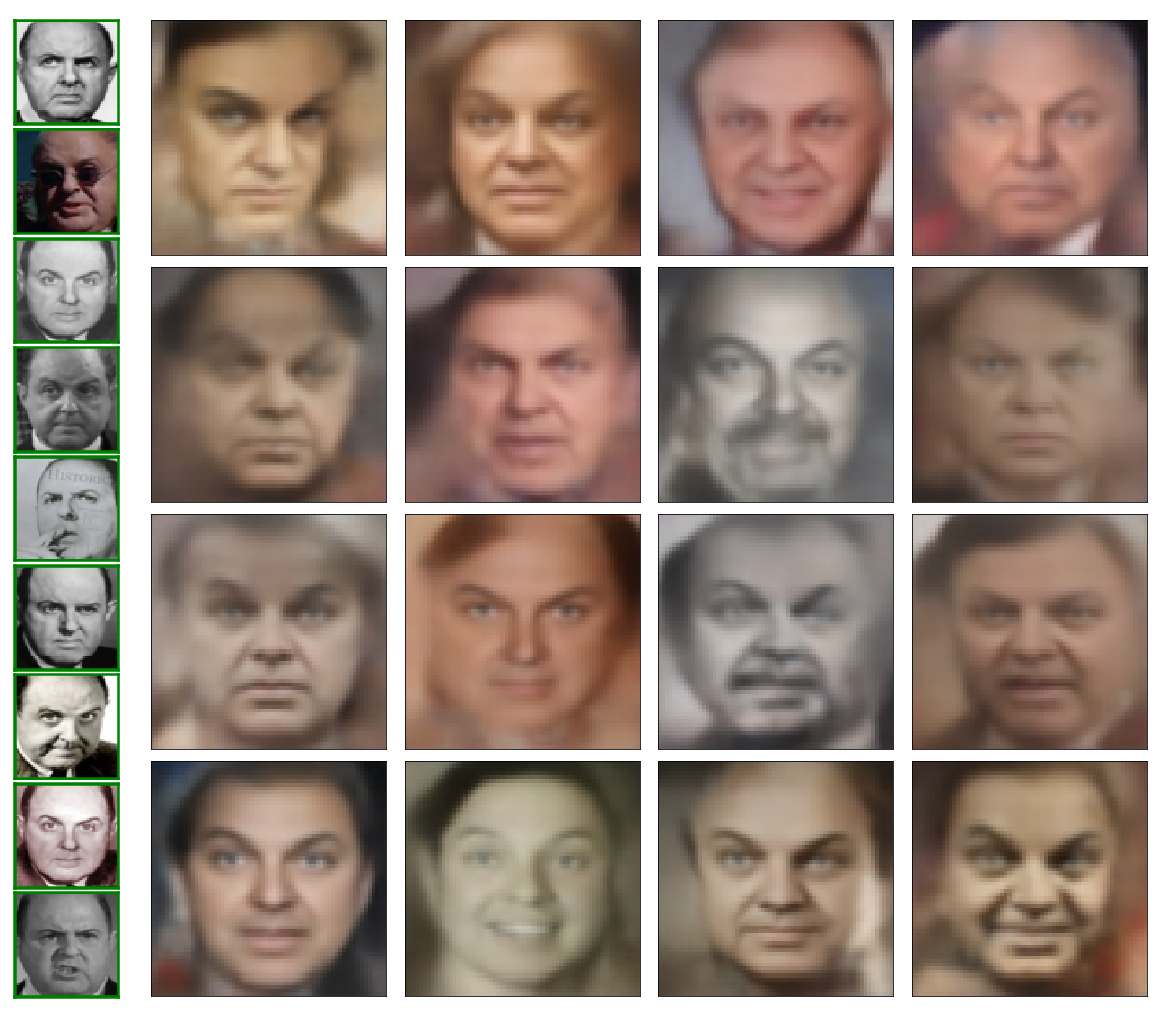}
    \mycaption{Generation, the green boxed images are all the test data images for this identity. On the right, sampling from the random prior for the style and using the mean of the grouped images latent code.}
\end{subfigure}
\hfill
\begin{subfigure}[t]{0.49\textwidth}
\centering
\captionsetup{width=.95\textwidth}
  \includegraphics[width=0.8\textwidth,trim={0cm 0cm 0cm 0cm},clip]{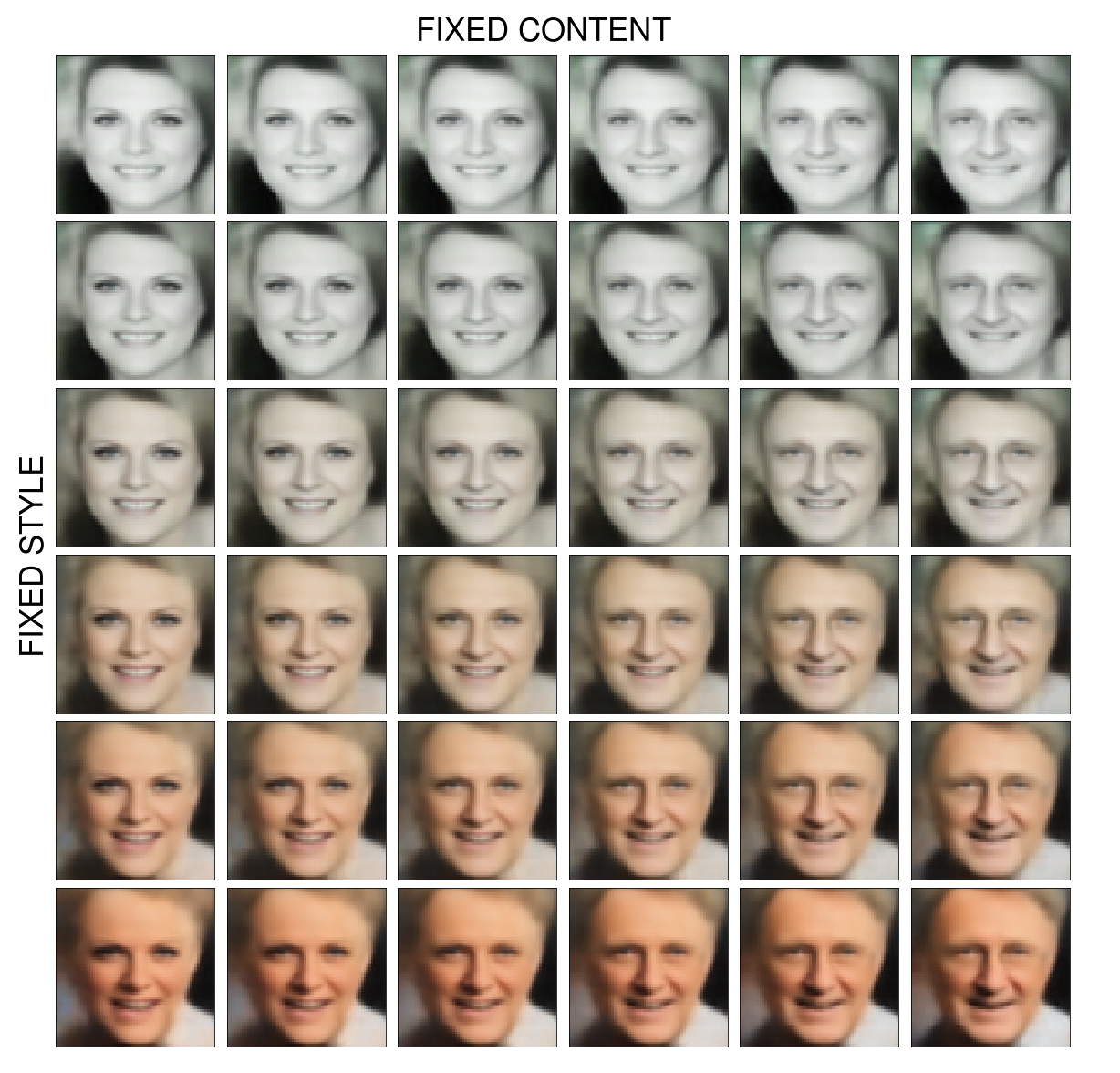}
  \mycaption{Interpolation, from top left to bottom right rows correspond to a fixed style and interpolating on the content, columns correspond to a fixed content and interpolating on the style.}
\end{subfigure}
\mycaption{Left: Generation. Right: Interpolation. Best viewed in color on screen.}
\label{interpolation}
\end{figure}
\begin{figure}[t]
\begin{minipage}[t]{.5\textwidth}
\begin{subfigure}[t]{0.49\textwidth}
\centering
\captionsetup{width=.95\textwidth}
  \includegraphics[width=0.8\textwidth, trim={1cm 2.5cm 1cm 0cm},clip]{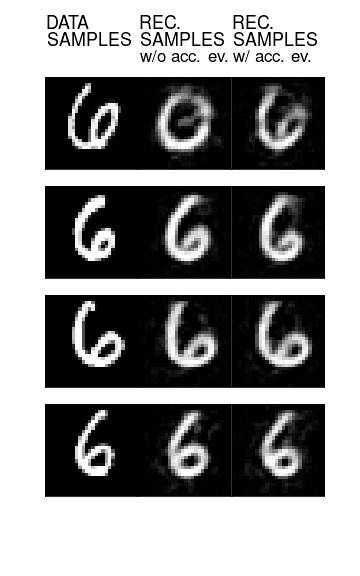}
  \mycaption{The four digits are of the same label.}
  \label{accumulation1}
\end{subfigure}
\begin{subfigure}[t]{0.49\textwidth}
\centering
\captionsetup{width=.95\textwidth}
  \includegraphics[width=0.85\textwidth, trim={1cm 3cm 1cm 0cm},clip]{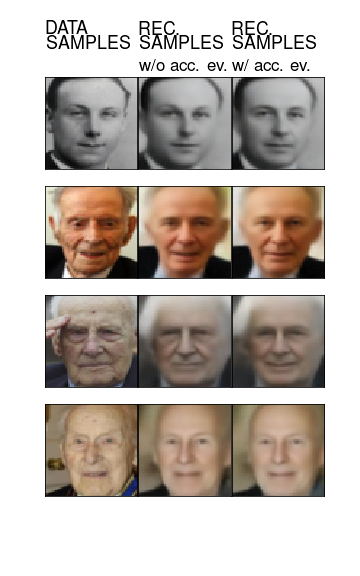}
  \mycaption{The four images are of the same person.}
  \label{accumulation2}
\end{subfigure}
\end{minipage}
\begin{minipage}[t]{.5\textwidth}
\begin{subfigure}[t]{\textwidth}
\centering
\captionsetup{width=.95\textwidth}
  \includegraphics[width=0.8\textwidth, trim={0cm 0cm 0cm 0cm},clip]{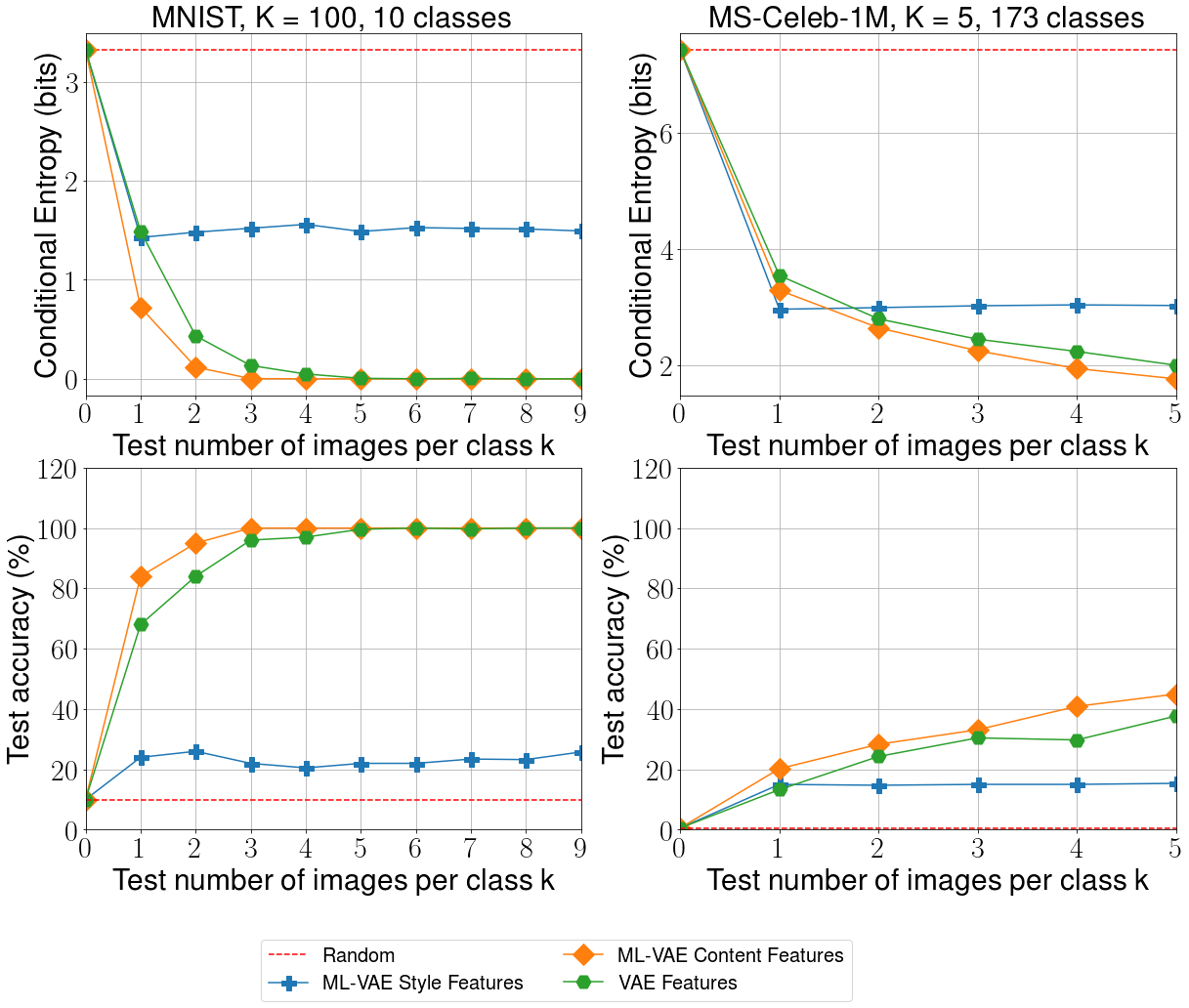}
  \mycaption{Quantitative Evaluation. For clarity on MNIST we show up to~$k=10$ as values stay stationary for larger~$k$ (in supplementary material).}
\label{quanttable}    
\end{subfigure}
\end{minipage}
\mycaption{Accumulating evidence (acc. ev.). Left column are test data samples, middle column are reconstructed sample without acc. ev., right column are reconstructed samples with acc. ev. from the four images. In (a), ML-VAE corrects inference (wrong digit label in first row second column) with acc. ev. (correct digit label in first row third column). In (b), where images of the same identity are taken at different ages, ML-VAE benefits from group information and the facial traits with acc. ev. (third column) are more constant than without acc. ev. (second column). Best viewed in color on screen.}
\vspace{-5mm}  
\end{figure}
Figure~\ref{swapping} shows the swapping procedure, where the first row and the first column show the test data sample input to ML-VAE, second row and column are reconstructed samples. Each row is a fixed style and each column is a fixed content. We see that the ML-VAE disentangles the factors of variation of the data in a relevant manner. In the case of MS-Celeb-1M, we see that the model encodes the factor of variation that grouped the data, that is the identity, into the facial traits which remain constant when we change the style, and encodes the style into the remaining factors (background color, face orientation for example). The ML-VAE learns this meaningful disentanglement without the knowledge that the images are grouped by identity, but only the organisation of the data into groups. Figure~\ref{interpolation} shows interpolation and generation. We see that our model covers the manifold of the data, and that style and content are disentangled. In Figures~\ref{accumulation1} and~\ref{accumulation2}, we reconstruct images of the same group with and without taking into account the grouping information. We see that the ML-VAE handles cases where there is no group information at test-time, and benefits from accumulating evidence if available. 
\vspace{-2mm}
\myparagraph{Quantitative Evaluation.}
In order to quantitatively evaluate the disentanglement power of ML-VAE, we use the style latent code~$S$ and content latent code~$C$ as features for a classification task. The quality of the disentanglement is high if the content~$C$ is informative about the class, while the style~$S$ is not. In the case of MNIST the class is the digit label and for MS-Celeb-1M the class is the identity. We emphasise that in the case of MS-Celeb-1M test images are all unseen classes (unseen identities) at training. We learn to classify the test images with a neural network classifier composed of two linear layers of~$256$ hidden units each, once using~$S$ and once using~$C$ as input features. Again we explore the benefits of accumulating evidence: while we construct the variational approximation on the content latent code by accumulating~$K$ images per class for training the classifier, we accumulate only~$k \le K$ images per class at test time, where~$k=1$ corresponds to no group information. When~$k$ increases we expect the performance of the classifer trained on~$C$ to improve as the features become more informative and the performance using features~$S$ to remain constant. We compare to the original VAE model, where we also accumulate evidence by using the Product of Normal method on the VAE latent code for samples of the same class. The results are shown in Figure~\ref{quanttable}. The ML-VAE content latent code is as informative about the class as the original VAE latent code, both in terms of classification accuracy and conditional entropy. ML-VAE also provides relevant disentanglement as the style remains uninformative about the class. Details on the choices of~$K$ and this experiment are in the supplementary material. 
\mysection{Discussion}
We proposed the Multi-Level VAE model for learning a meaningful disentanglement from a set of grouped observations. The ML-VAE model handles an arbitrary number of groups of observations, which needs not be the same at training and testing. We proposed different methods for incorporating the semantics embedded in the grouping. Experimental evaluation show the relevance of our method, as the ML-VAE learns a semantically meaningful disentanglement, generalises to unseen groups and enables control on the latent representation. For future work, we wish to apply the ML-VAE to text data.
\newpage
\bibliographystyle{plainnat} 
\bibliography{refs} 

\begin{thebibliography}{35}
\providecommand{\natexlab}[1]{#1}
\providecommand{\url}[1]{\texttt{#1}}
\expandafter\ifx\csname urlstyle\endcsname\relax
  \providecommand{\doi}[1]{doi: #1}\else
  \providecommand{\doi}{doi: \begingroup \urlstyle{rm}\Url}\fi

\bibitem[Abbasnejad et~al.(2016)Abbasnejad, Dick, and van~den
  Hengel]{Abbasnejad2016}
Ehsan Abbasnejad, Anthony~R. Dick, and Anton van~den Hengel.
\newblock Infinite variational autoencoder for semi-supervised learning.
\newblock \emph{arXiv preprint arXiv:1611.07800}, 2016.

\bibitem[Alemi et~al.(2017)Alemi, Fischer, Dillon, and Murphy]{Alemi2017}
Alexander~A. Alemi, Ian Fischer, Joshua~V. Dillon, and Kevin Murphy.
\newblock Deep variational information bottleneck.
\newblock \emph{ICLR}, 2017.

\bibitem[Allamanis et~al.(2017)Allamanis, Chanthirasegaran, Kohli, and
  Sutton]{Allamanis2017}
Miltiadis Allamanis, Pankajan Chanthirasegaran, Pushmeet Kohli, and Charles
  Sutton.
\newblock Learning continuous semantic representations of symbolic expressions.
\newblock \emph{arXiv preprint 1611.01423}, 2017.

\bibitem[Bengio et~al.(2013)Bengio, Courville, and Vincent]{Bengio:2013}
Yoshua Bengio, Aaron Courville, and Pascal Vincent.
\newblock Representation learning: A review and new perspectives.
\newblock \emph{IEEE Trans. Pattern Anal. Mach. Intell.}, 35\penalty0
  (8):\penalty0 1798--1828, August 2013.
\newblock ISSN 0162-8828.

\bibitem[Bouchacourt et~al.(2016)Bouchacourt, Mudigonda, and
  Nowozin]{Bouchacourt2016}
Diane Bouchacourt, Pawan~Kumar Mudigonda, and Sebastian Nowozin.
\newblock {DISCO} nets : Dissimilarity coefficients networks.
\newblock \emph{NIPS}, 2016.

\bibitem[Bousmalis et~al.(2016)Bousmalis, Silberman, Dohan, Erhan, and
  Krishnan]{Bousmalis2016}
Konstantinos Bousmalis, Nathan Silberman, David Dohan, Dumitru Erhan, and Dilip
  Krishnan.
\newblock Unsupervised pixel-level domain adaptation with generative
  adversarial networks.
\newblock \emph{arXiv preprint arXiv:1612.05424}, 2016.

\bibitem[Chen et~al.(2016)Chen, Duan, Houthooft, Schulman, Sutskever, and
  Abbeel]{Chen2016}
Xi~Chen, Yan Duan, Rein Houthooft, John Schulman, Ilya Sutskever, and Pieter
  Abbeel.
\newblock Info{GAN}: Interpretable representation learning by information
  maximizing generative adversarial nets.
\newblock \emph{NIPS}, 2016.

\bibitem[Chen et~al.(2017)Chen, Kingma, Salimans, Duan, Dhariwal, Schulman,
  Sutskever, and Abbeel]{Chen2017}
Xi~Chen, Diederik~P. Kingma, Tim Salimans, Yan Duan, Prafulla Dhariwal, John
  Schulman, Ilya Sutskever, and Pieter Abbeel.
\newblock Variational lossy autoencoder.
\newblock \emph{ICLR}, 2017.

\bibitem[Donahue et~al.(2017)Donahue, Balsubramani, McAuley, and
  Lipton]{Donahue2017}
Chris Donahue, Akshay Balsubramani, Julian McAuley, and Zachary~C. Lipton.
\newblock Semantically decomposing the latent spaces of generative adversarial
  networks.
\newblock \emph{arXiv preprint 1705.07904}, 2017.

\bibitem[Dziugaite et~al.(2015)Dziugaite, Roy, and Ghahramani]{Dziugaite2015}
Gintare~Karolina Dziugaite, Daniel~M. Roy, and Zoubin Ghahramani.
\newblock Training generative neural networks via maximum mean discrepancy
  optimization.
\newblock \emph{UAI}, 2015.

\bibitem[Edwards and Storkey(2015)]{Edwards2016}
Harrison Edwards and Amos~J. Storkey.
\newblock Censoring representations with an adversary.
\newblock \emph{CoRR}, 2015.

\bibitem[{Fu} et~al.(2017){Fu}, {Liu}, {Chiu}, {Wang}, and {Wang}]{Fu2017}
T.-C. {Fu}, Y.-C. {Liu}, W.-C. {Chiu}, S.-D. {Wang}, and Y.-C.~F. {Wang}.
\newblock {Learning Cross-Domain Disentangled Deep Representation with
  Supervision from A Single Domain}.
\newblock \emph{arXiv preprint arXiv:1705.01314}, 2017.

\bibitem[Goodfellow et~al.(2014)Goodfellow, Pouget-Abadie, Mirza, Xu,
  Warde-Farley, Ozair, Courville, and Bengio]{Goodfellow2014}
Ian Goodfellow, Jean Pouget-Abadie, Mehdi Mirza, Bing Xu, David Warde-Farley,
  Sherjil Ozair, Aaron Courville, and Yoshua Bengio.
\newblock Generative adversarial nets.
\newblock \emph{NIPS}, 2014.

\bibitem[Guo et~al.(2016)Guo, Zhang, Hu, He, and Gao]{Guo2016}
Yandong Guo, Lei Zhang, Yuxiao Hu, Xiaodong He, and Jianfeng Gao.
\newblock M{S}-{C}eleb-1{M}: A dataset and benchmark for large scale face
  recognition.
\newblock \emph{ECCV}, 2016.

\bibitem[Higgins et~al.(2017)Higgins, Matthey, Pal, Burgess, Glorot, Botvinick,
  Mohamed, and Lerchner]{Higgins2017}
Irina Higgins, Loic Matthey, Arka Pal, Christopher Burgess, Xavier Glorot,
  Matthew Botvinick, Shakir Mohamed, and Alexander Lerchner.
\newblock beta-{VAE}: Learning basic visual concepts with a constrained
  variational framework.
\newblock \emph{ICLR}, 2017.

\bibitem[Hoffman et~al.(2013)Hoffman, Blei, Wang, and Paisley]{Hoffman2013}
Matthew~D. Hoffman, David~M. Blei, Chong Wang, and John Paisley.
\newblock Stochastic variational inference.
\newblock \emph{JMLR}, 2013.

\bibitem[Kim et~al.(2017)Kim, Cha, Kim, Lee, and Kim]{Kim2017}
T~Kim, M~Cha, H~Kim, J~Lee, and J~Kim.
\newblock Learning to discover cross-domain relations with generative
  adversarial networks.
\newblock \emph{arXiv preprint arXiv:1703.05192}, 2017.

\bibitem[Kingma and Welling(2014)]{Kingma2013}
Diederik~P. Kingma and Max Welling.
\newblock {A}uto-{E}ncoding {V}ariational {B}ayes.
\newblock \emph{ICLR}, 2014.

\bibitem[Kulkarni et~al.(2015)Kulkarni, Whitney, Kohli, and
  Tenenbaum]{Kulkarni2015}
Tejas~D Kulkarni, Will Whitney, Pushmeet Kohli, and Joshua~B Tenenbaum.
\newblock Deep convolutional inverse graphics network.
\newblock \emph{NIPS}, 2015.

\bibitem[Lecun et~al.(1998)Lecun, Bottou, Bengio, and Haffner]{Lecun1998}
Yann Lecun, Léon Bottou, Yoshua Bengio, and Patrick Haffner.
\newblock Gradient-based learning applied to document recognition.
\newblock \emph{Proceedings of the IEEE}, pages 2278--2324, 1998.

\bibitem[Li et~al.(2015)Li, Swersky, and Zemel]{Li2015}
Yujia Li, Kevin Swersky, and Richard~S. Zemel.
\newblock Generative moment matching networks.
\newblock \emph{ICML}, 2015.

\bibitem[Linsker(1988)]{Linsker1998}
Ralph Linsker.
\newblock Self-organization in a perceptual network.
\newblock \emph{Computer}, 21\penalty0 (3):\penalty0 105--117, 1988.

\bibitem[Liu et~al.(2017)Liu, Breuel, and Kautz]{Liu2017}
Ming{-}Yu Liu, Thomas Breuel, and Jan Kautz.
\newblock Unsupervised image-to-image translation networks.
\newblock \emph{arXiv preprint arXiv:1703.00848}, 2017.

\bibitem[Louizos et~al.(2016)Louizos, Swersky, Li, Welling, and
  Zemel]{Louizos2016}
Christos Louizos, Kevin Swersky, Yujia Li, Max Welling, and Richard~S. Zemel.
\newblock The variational fair autoencoder.
\newblock \emph{ICLR}, 2016.

\bibitem[Mathieu et~al.(2016)Mathieu, Zhao, Zhao, Ramesh, Sprechmann, and
  LeCun]{Mathieu2016}
Michael~F Mathieu, Junbo~Jake Zhao, Junbo Zhao, Aditya Ramesh, Pablo
  Sprechmann, and Yann LeCun.
\newblock Disentangling factors of variation in deep representation using
  adversarial training.
\newblock \emph{NIPS}, 2016.

\bibitem[Murali et~al.(2017)Murali, Chaudhuri, and Jermaine]{Murali2017}
Vijayaraghavan Murali, Swarat Chaudhuri, and Chris Jermaine.
\newblock Bayesian sketch learning for program synthesis.
\newblock \emph{arXiv preprint arXiv:1703.05698v2}, 2017.

\bibitem[Murphy(2007)]{Murphy2007}
Kevin~P. Murphy.
\newblock Conjugate {B}ayesian {A}nalysis of the {G}aussian {D}istribution.
\newblock Technical report, 2007.

\bibitem[Rezende et~al.(2014)Rezende, Mohamed, and Wierstra]{Rezende2014}
Danilo~Jimenez Rezende, Shakir Mohamed, and Daan Wierstra.
\newblock Stochastic backpropagation and approximate inference in deep
  generative models.
\newblock \emph{ICML}, 2014.

\bibitem[Shrivastava et~al.(2017)Shrivastava, Pfister, Tuzel, Susskind, Wang,
  and Webb]{Shrivastava2016}
Ashish Shrivastava, Tomas Pfister, Oncel Tuzel, Josh Susskind, Wenda Wang, and
  Russ Webb.
\newblock Learning from simulated and unsupervised images through adversarial
  training.
\newblock \emph{arXiv preprint arXiv:1612.07828}, 2017.

\bibitem[Siddharth et~al.(2017)Siddharth, Paige, Desmaison, Wood, and
  Torr]{Siddharth2017}
N.~Siddharth, Brooks Paige, Alban Desmaison, Frank Wood, and Philip Torr.
\newblock Learning disentangled representations in deep generative models.
\newblock \emph{Submitted to ICLR}, 2017.

\bibitem[Taigman et~al.(2017)Taigman, Polyak, and Wolf]{Taigman2016}
Yaniv Taigman, Adam Polyak, and Lior Wolf.
\newblock Unsupervised cross-domain image generation.
\newblock \emph{ICLR}, 2017.

\bibitem[Tishby et~al.(1999)Tishby, Pereira, and Bialek]{Tishby1999}
N.~Tishby, F.~C. Pereira, and W.~Bialek.
\newblock The information bottleneck method.
\newblock \emph{37th annual Allerton Conference on Communication, Control and
  Computing}, 1999.

\bibitem[Wang and Gupta(2016)]{Wang_SSGAN2016}
Xiaolong Wang and Abhinav Gupta.
\newblock Generative image modeling using style and structure adversarial
  networks.
\newblock \emph{ECCV}, 2016.

\bibitem[Yi et~al.(2017)Yi, Zhang, Tan, and Gong]{Yi2017}
Zili Yi, Hao Zhang, Ping Tan, and Minglun Gong.
\newblock Dualgan: Unsupervised dual learning for image-to-image translation.
\newblock \emph{arXiv preprint arXiv:1704.02510}, 2017.

\bibitem[Zhu et~al.(2017)Zhu, Park, Isola, and Efros]{Zhu2017}
Jun-Yan Zhu, Taesung Park, Phillip Isola, and Alexei~A Efros.
\newblock Unpaired image-to-image translation using cycle-consistent
  adversarial networks.
\newblock \emph{arXiv preprint arXiv:1703.10593}, 2017.

\end{thebibliography}



\section*{Supplementary material}

# Multi-Level Variational Autoencoder: Learning Disentangled Representations from Grouped Observations
# Supplementary Material

**Diane Bouchacourt**
OVAL Group
University of Oxford*
diane@robots.ox.ac.uk

**Ryota Tomioka, Sebastian Nowozin**
Machine Intelligence and Perception Group
Microsoft Research
Cambridge, UK
{ryoto,Sebastian.Nowozin}@microsoft.com

## 1 Mixture of Normals Method

We discuss here a method to construct the variational approximation $q(C_G = c|\boldsymbol{X_G} = \boldsymbol{x}_G, \phi_c)$ as a mixture of $|G|$ density functions, each of them evaluating the probability of $C_G = c$ given $X_i = x_i$. This is an alternative to the Product of Normals method.

$$q(C_G = c|\boldsymbol{X_G} = \boldsymbol{x}_G, \phi_c) = \frac{1}{|G|} \sum_{i=1}^{|G|} q(C_G = c|X_i = x_i, \phi_c) \tag{1}$$

We assume $q(C_G|X_i = x_i, \phi_c)$ to be a Normal distribution $N(\mu_i, \Sigma_i)$. However, the term $\mathrm{KL}(q(C_G|\mathbf{X}_G; \phi_c)||p(C_G))$ can not be computed in closed form

$$\begin{aligned}\mathrm{KL}(q(C_G|\mathbf{X}_G; \phi_c)||p(C_G)) &= \mathbb{E}_{q(C_G|\boldsymbol{X}_G, \phi_c)}[\log q(C_G|\mathbf{X}_G, \phi_c) - \log p(C_G)] \\ &= \mathbb{E}_{q(C_G|\boldsymbol{X}_G, \phi_c)}[\log q(C_G|\mathbf{X}_G, \phi_c)] - \mathbb{E}_{q(C_G|\boldsymbol{X}_G, \phi_c)}[\log p(C_G)]\end{aligned} \tag{2}$$

We estimate this term by sampling $L$ samples $c_l \sim q(C_G|\mathbf{X}_G; \phi_c)$ and computing the estimate:

$$\frac{1}{L}\sum_{l=1}^{L} \log \frac{1}{|G|} \sum_{i=1}^{|G|} q(C_G = c_l|X_i = x_i, \phi_c) - \frac{1}{L}\sum_{l=1}^{L} \log p(C_G = c_l) \tag{3}$$

In our experiments we used $L = |G|$ as we used the samples we draw to compute the first term of the objective function, that is $\sum_{i \in G} \mathbb{E}_{q(C_G|\mathbf{X}_G;\phi_c)}[\mathbb{E}_{q(S_i|X_i;\phi_s)}[\log p(X_i|C_G, S_i; \theta)]]$.

Figures 1 shows the qualitative results of the Mixture of Normal method. Qualitative evaluation indicate a better disentanglement with the Product of Normal method therefore we focused on the Product. On MS-Celeb-1M, the Mixture of Normal densities seems to store information about the grouping into the style: when style gets transfered, the facial features along the columns (which should remain constant as it is the same identity) tend to change[2]. Nevertheless, we present it as it might be better suited to other datasets and other tasks.

## 2 Experimental Details

In all experiments we use the Adam optimiser presented in Kingma and Ba [2015] with $\alpha = 0.001, \beta_1 = 0.9, \beta_2 = 0.999, \epsilon = 1e - 8$. We use diagonal covariances for the Normal variational

---

*The work was performed as part of an internship at Microsoft Research.

[2]The previous version of the supplementary material had the wrong figure on MS-Celeb-1M (computed with another model). This updated figure shows the actual results and emphasize the conclusion that information about content is stored in the style.

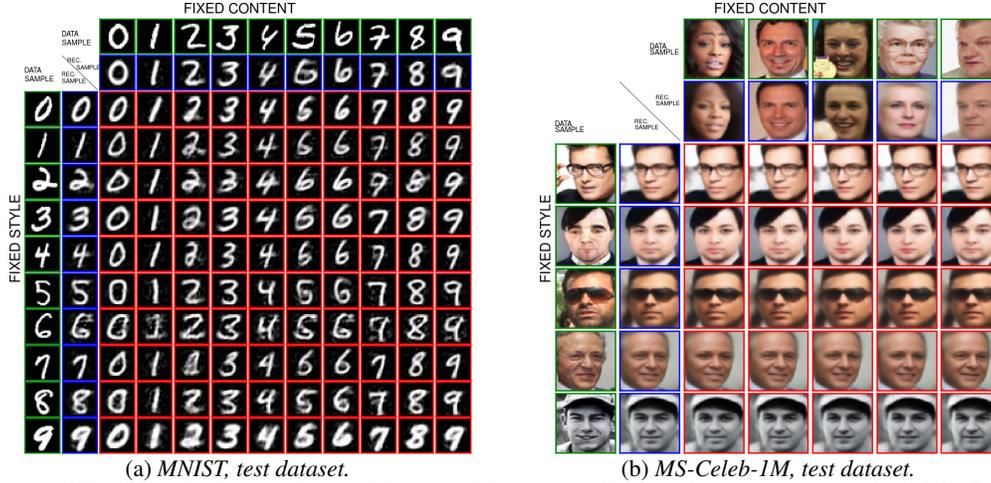

(a) *MNIST, test dataset.*  (b) *MS-Celeb-1M, test dataset.*

Figure 1: *ML-VAE with Mixture of Normal densities. We intentionally show the same images as the ML-VAE with Product of Normal for comparison purposes. Swapping, first row and first column are test data samples (green boxes), second row and column are reconstructed samples (blue boxes) and the rest are swapped reconstructed samples (red boxes). Each row is fixed style and each column is a fixed content. Best viewed in color on screen. As we mention, compared to ML-VAE with Product of Normal, the Mixture of Normal seems to store information about the grouping into the style.*

approximations $q(\mathbf{S}_G|\mathbf{X}_G; \phi_s), q(C_G|\mathbf{X}_G; \phi_c)$. in both experiments we use a Normal distribution with diagonal covariances for the posterior $p(X_i|C_G, S_i; \theta)$. We train the model for 2000 epochs on MNIST and 250 epochs on MS-Celeb-1M. When we compare the original VAE and the ML-VAE we use early stopping on the validation set for MS-Celeb-1M as the architecture is larger and prone to over-fitting. In the specific case of Stochastic Variational Inference (SVI) Hoffman et al. [2013], for MNIST we train the model for 2000 epochs and proceed to 40000 epochs of inference at test-time, and for MS-Celeb-1M we use 500 training epochs and proceed to 200 epochs of inference at test-time.

## 2.1 Networks Architectures

**MNIST Lecun et al. [1998].** We use an encoder network composed of a first linear layer $e_0$ that takes as input a $1 \times 784$-dimensional MNIST image $x_i$, $x_i$ is a realisation of $X_i$. Layer $e_0$ has 500 hidden units and the hyperbolic tangent activation function. After layer $e_0$ we separate the network into 4 linear layers, $e_{m,s}, e_{v,s}$ and $,e_{m,c}, e_{v,c}$ each of size $500 \times d$ where $d$ is the dimension of both latent representations $S$ and $C$. The layers $e_{m,s}, e_{v,s}$ take as input the output of $e_0$ and output respectively the mean and log-variance of the Normal distribution $q(S_i|X_i = x_i; \phi_s)$. The layers $e_{m,c}, e_{v,c}$ take as input the output of $l_0$ and output respectively the mean and variance of the Normal distribution $q(C_G|X_i = x_i; \phi_c)$.

We then construct $q(C_G|\mathbf{X}_G; \phi_c)$ from $q(C_G|X_i = x_i; \phi_c), i \in G$. Let us denote $G$ the group in which $x_i$ belongs. As explained in step 9 of Algorithm 1 in the main paper, for each input $x_i$ we draw a sample $c_{G,i} \sim q(C_G|\mathbf{X}_G = \mathbf{x}_G; \phi_c)$ for the content of the group $G$, and a sample $s_i \sim q(S_i|X_i = x_i; \phi_s)$ of the style latent representation. We concatenate $(c_{G,i}, s_i)$ into a $2 \times d$-dimensional vector that is fed to the decoder.

The decoder network is composed of a first linear layer $d_0$ that takes as input the $2 \times d$-dimensional vector $(c_{G,i}, s_i)$. Layer $d_0$ has 500 hidden units and the hyperbolic tangent activation function. A second linear layer $d_2$ takes as input the output of $d_0$ and outputs a 784-dimensional vector representating the parameters of the Normal distribution $p(X_i|C_G, S_i; \theta)$. We use in our experiments $d = 10$ for a total latent representation size of respectively 20.

**MS-Celeb-1M Guo et al. [2016].** We use an encoder network composed of a four convolutional layers $e_1, e_2, e_3, e_4$ all of stride 2 and kernel size 4. They are composed of respectively 64, 128, 256and512 filters. All four layers are followed by Batch Normalisation and Rectified Linear Units (ReLU) activation functions. The fifth and sixth layers $e_5, e_6$ are linear layers with 256 hidden units, followed by Batch Normalisation and Concatenated Rectified Linear Unit (CReLU) activation functions. Similarly to the MNIST architecture, after layer $e_6$ we separate the network

into 4 linear layers, $e_{m,s}, e_{v,s}$ and , $e_{m,c}, e_{v,c}$ each of size $256 \times 2 \times d$ where $d$ is the dimension of both latent representations $S$ and $C$. The layers for the log-variances are followed by the tangent hyperbolic activation function and multiplied by 5.

Similarly as the MNIST experiment we construct the latent representation and sample it as explained in Algorithm 1 in the main paper.

The decoder network is composed of 3 deconvolutional layers $d_1, d_2, d_3$ all of stride 2 and kernel size 4. They are composed of respectively $256, 128, 64$ filters. All four layers are followed by Batch Normalisation and Rectified Linear Units (ReLU) activation functions. The seventh and eight layers are deconvolutional layers composed of 3 filters, of stride 1 and kernel size 3 and output respectively the mean and log-variance of the Normal distribution $p(X_i|C_G, S_i; \theta)$. The layer for the log-variances are followed by the tangent hyperbolic activation function and multiplied by 5. We use in our experiments $d = 50$ for a total latent representation size 100. We use padding in the convolutional and deconvolutional layers to match the data size.

**Specific case for Stochastic Variational Inference (SVI) Hoffman et al. [2013].** We compare in our experiments with Stochastic Variational Inference (SVI), from Hoffman et al. [2013]. In the case of SVI, the encoder is an embedding layer mapping each sample $x_i$ in a group $G$ to the non-shared parameters $\phi_{s,i}$ of its style latent representation $q(S_i|\phi_{s,i})$ and to the non-shared parameters $\phi_{c,G}$ of its group content latent representation $q(C_G|\phi_{c,G})$. The decoder is the same as the ML-VAE.

## 3 Quantitative Evaluation details

We explain in Section 4 of the main paper how we quantitatively evaluate the disentanglement performance of our model. We give details here for the interested reader. Figure 2 show the quantitative evaluation for $k$ up to $k = K = 100$ on MNIST.

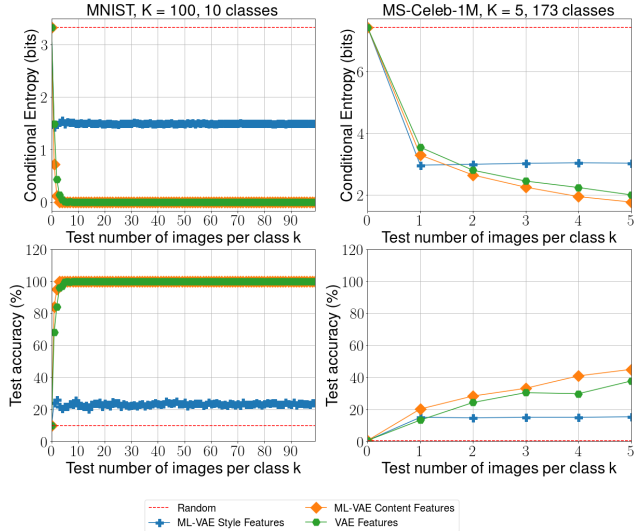

Figure 2: *Quantitative Evaluation.*

### 3.1 Classifier architecture

The classifier is a neural network composed of two linear layers of 256 hidden units each. The first layer is followd by a tangent hyperbolic activation function. The second layer is followed by a softmax activation function. We use the cross-entropy loss. We use the Adam optimiser presented in Kingma and Ba [2015] with $\alpha = 0.001, \beta_1 = 0.9, \beta_2 = 0.999, \epsilon = 1e - 8$. Note that the training, validation and testing sets for the classifier are all composed of test images, and each set is composed of $K$ times the number of classes; hence our choice of $K$ and number of classes for MS-Celeb-1M. In the case of MNIST, there are only 10 classes, therefore when $k$ is small we would take only a small number of images to test the classifier. Therefore we perform 5 trials of test procedures of the classifier, each trial using different test images, and report the mean performance.

## 3.2 Conditional entropy computation.

We show here that training the neural network classifier with the cross-entropy loss is a proxy for minimising the conditional entropy of the class given the latent code features $C$ or $S$.

Let us take the example of the latent code $C$ used as features. We denote a class $Y$ and we train the neural network classifier to model the distribution $r(Y|C)$ by minimising the cross-entropy loss, which corresponds to maximising $\mathbb{E}_{p(Y,C)}[\log r(Y|C)]$ where $p(Y,C)$ is estimated using samples

$$
\begin{aligned}
&\text{Sample a class } y \sim p(Y), \\
&\text{Sample grouped observations for this class } \mathbf{x}_{G_Y} \sim p(\mathbf{X}_{G_Y}), \\
&\text{Sample the latent code to use as features } c_{G_Y} \sim q(C_{G_Y}|\mathbf{X}_{G_Y}, \phi_c)
\end{aligned}
\quad (4)
$$

The conditional entropy of the class $Y$ given the latent code $C$ is expressed as

$$\mathbb{H}(Y|C) = -\mathbb{E}_{p(Y,C)}[\log p(Y|C)] \quad (5)$$

We can write

$$
\begin{aligned}
\mathbb{H}(Y|C) &= -\mathbb{E}_{p(Y,C)}[\log p(Y|C)] = -\mathbb{E}_{p(Y,C)}[\log \frac{p(Y|C)}{r(Y|C)} r(Y|C)] \\
&= -\mathbb{E}_{p(Y,C)}[\log r(Y|C)] - \mathbb{E}_{p(Y,C)}[\log \frac{p(Y|C)}{r(Y|C)}] \\
&= -\mathbb{E}_{p(Y,C)}[\log r(Y|C)] - \mathbb{E}_{p(Y,C)}[\log \frac{p(Y,C)}{r(Y,C)}] \\
&= -\mathbb{E}_{p(Y,C)}[\log r(Y|C)] - \text{KL}(p(Y,C)||r(Y,C)) \leq -\mathbb{E}_{p(Y,C)}[\log r(Y|C)]
\end{aligned}
\quad (6)
$$

since the Kullback-Leibler is always positive. Therefore, training the neural network classifier to minimise the cross-entropy loss is equivalent to minimising an upper bound on the conditional entropy of the class given the latent code features $C$. We report the value of $\mathbb{E}_{p(Y,C)}[\log r(Y|C)]$ on the classifier test set in the paper as the reported Conditional entropy in bits. Similarly we report the performances with the style latent code or the latent code of the original VAE model.

## 4 ML-VAE with Product of Normal Without Accumulating Evidence

We show in Figure 3 and 4 the results of swapping and interpolation of the ML-VAE with Product of Normal without accumulating evidence (strategy 1 in the main paper). We intentionally show the same images for comparison purposes.

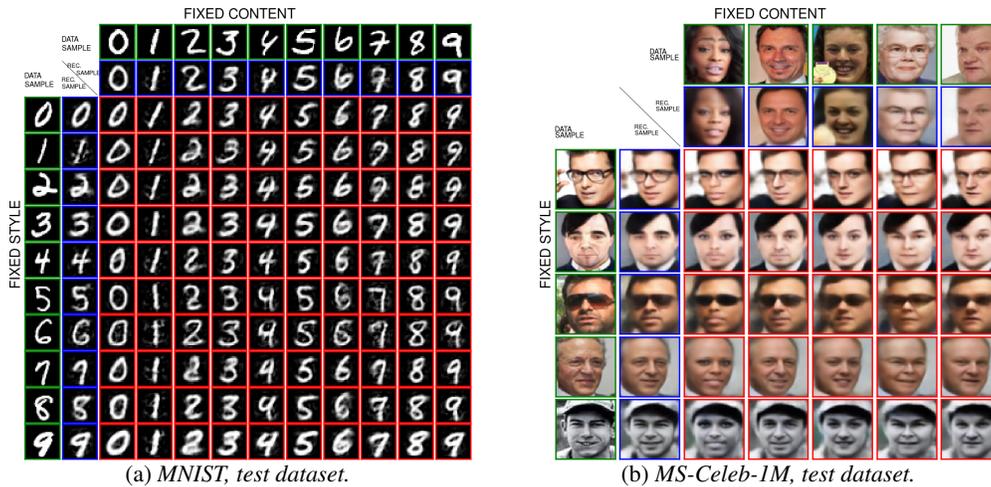

(a) *MNIST, test dataset.*  (b) *MS-Celeb-1M, test dataset.*

Figure 3: *ML-VAE with Product of Normal without accumulating evidence. Swapping, first row and first column are test data samples (green boxes), second row and column are reconstructed samples (blue boxes) and the rest are swapped reconstructed samples (red boxes). Each row is fixed style and each column is a fixed content. Best viewed in color on screen.*

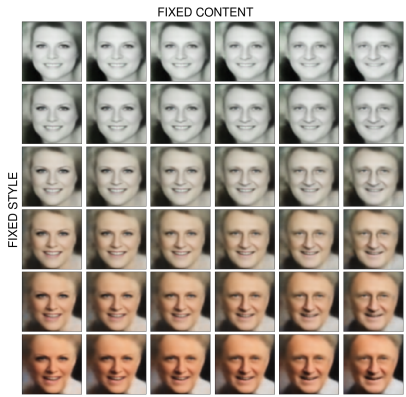

Figure 4: *ML-VAE without accumulating evidence. Interpolation, from top left to bottom right rows correspond to a fixed style and interpolating on the content, columns correspond to a fixed content and interpolating on the style.*

# 5 Accumulating Group Evidence using a Product of Normal densities: Detailed derivations

We construct the probability density function of the random variable $C_G$ by multiplying $|G|$ normal density functions, each of them evaluating the probability of $C_G$ under the realisation $X_i = x_i$, where $i \in G$.

$$q(C_G|\mathbf{X}_G = \mathbf{x}_G; \phi_c) \propto \prod_{i \in G} q(C_G|X_i = x_i; \phi_c) \tag{7}$$

We assume $q(C_G|X_i = x_i; \phi_c)$ to be a Normal distribution $N(\mu_i, \Sigma_i)$. The normalisation constant is the resulting product marginalised over all possible values of $C_G$.

The result of the product of $|G|$ normal density functions is proportional to the density function of a Normal distribution of mean $\mu_G$ and variance $\Sigma_G$.

$$\begin{aligned} \Sigma_G^{-1} &= \sum_{i \in G} \Sigma_i^{-1} \\ \mu_G^T \Sigma_G^{-1} &= \sum_{i \in G} \mu_i^T \Sigma_i^{-1} \end{aligned} \tag{8}$$

We show below how we derive the expressions of mean $\mu_G$ and variance $\Sigma_G$.

$$
\begin{aligned}
\prod_{i \in G} q(C_G = c | X_i = x_i; \phi_c) &= \prod_{i \in G} \frac{1}{\sqrt{(2\pi)^d |\Sigma_i|}} \exp\left(-\frac{1}{2}(c - \mu_i)^T \Sigma_i^{-1}(c - \mu_i)\right) \\
&= K_1 \exp\left(-\frac{1}{2} \sum_{i \in G}(c - \mu_i)^T \Sigma_i^{-1}(c - \mu_i)\right) \\
&= K_1 \exp\left(-\frac{1}{2} \Big(\sum_{i \in G} c^T \Sigma_i^{-1} c + \mu_i^T \Sigma_i \mu_i - 2\mu_i^T \Sigma_i^{-1} c\Big)\right) \\
&= K_1 K_2 \exp\left(\sum_{i \in G} \mu_i^T \Sigma_i^{-1} c - \frac{1}{2} c^T \Sigma_i^{-1} c\right) \\
&= K_1 K_2 \exp\left(\sum_{i \in G} \mu_i^T \Sigma_i^{-1} c - \frac{1}{2} c^T \sum_{i \in G} \Sigma_i^{-1} c\right) \\
&= K_1 K_2 \exp\left(\mu_G^T \Sigma_G^{-1} c - \frac{1}{2} c^T \Sigma_G^{-1} c\right) \\
&= K_1 K_2 \exp\left(-\frac{1}{2}(c^T \Sigma_G^{-1} c - 2\mu_G^T \Sigma_G^{-1} c)\right) \\
&= K_1 K_2 \exp\left(-\frac{1}{2}(c^T \Sigma_G^{-1} c - 2\mu_G^T \Sigma_G^{-1} c + \mu_G^T \Sigma_G^{-1} \mu_G - \mu_G^T \Sigma_G^{-1} \mu_G)\right) \\
&= K_1 K_2 \exp(-\mu_G^T \Sigma_G^{-1} \mu_G) \exp\left(-\frac{1}{2}(c^T \Sigma_G^{-1} c - 2\mu_G^T \Sigma_G^{-1} c + \mu_G^T \Sigma_G^{-1} \mu_G)\right) \\
&= K_1 K_2 K_3 \exp\left(-\frac{1}{2}(c^T \Sigma_G^{-1} c - 2\mu_G^T \Sigma_G^{-1} c + \mu_G^T \Sigma_G^{-1} \mu_G)\right) \\
&= K_1 K_2 K_3 K_4 \frac{1}{\sqrt{(2\pi)^d |\Sigma_G|}} \exp\left(-\frac{1}{2}(c - \mu_G)^T \Sigma_G^{-1}(c - \mu_G)\right)
\end{aligned}
\tag{9}
$$

where $d$ is the dimensionality of $c$ and

$$
\begin{aligned}
K_1 &= \prod_{i \in G} \frac{1}{\sqrt{(2\pi)^d |\Sigma_i|}} \\
K_2 &= \exp\left(-\frac{1}{2} \sum_{i \in G} \mu_i^T \Sigma_i^{-1} \mu_i\right) \\
K_3 &= \exp\left(\frac{1}{2} \mu_G^T \Sigma_G^{-1} \mu_G\right) \\
K_4 &= \sqrt{(2\pi)^d |\Sigma_G|}
\end{aligned}
\tag{10}
$$

This is a normal distribution, scaled by $K_1 K_2 K_3 K_4$, of mean $\mu_G$ and variance $\Sigma_G$

$$
\begin{aligned}
\Sigma_G^{-1} &= \sum_{i \in G} \Sigma_i^{-1}, \\
\mu_G^T \Sigma_G^{-1} &= \sum_{i \in G} \mu_i^T \Sigma_i^{-1}
\end{aligned}
\tag{11}
$$

However, the constant of normalisation disappears when we rescale the resulting product in order for the resulting product to integrate to 1.

$$q(C_G = c | X_G = x_G; \phi_c) = \frac{K_1 K_2 K_3 K_4 \frac{1}{\sqrt{(2\pi)^d |\Sigma_G|}} \exp\left(-\frac{1}{2}(c - \mu_G)^T \Sigma_G^{-1}(c - \mu_G)\right)}{\int_c K_1 K_2 K_3 K_4 \frac{1}{\sqrt{(2\pi)^d |\Sigma_G|}} \exp\left(-\frac{1}{2}(c - \mu_G)^T \Sigma_G^{-1}(c - \mu_G)\right) dc}$$

$$= \frac{\exp\left(-\frac{1}{2}(c - \mu_G)^T \Sigma_G^{-1}(c - \mu_G)\right)}{\int_c \exp\left(-\frac{1}{2}(c - \mu_G)^T \Sigma_G^{-1}(c - \mu_G)\right) dc}$$

$$= \frac{1}{\sqrt{(2\pi)^d |\Sigma_G|}} \exp\left(-\frac{1}{2}(c - \mu_G)^T \Sigma_G^{-1}(c - \mu_G)\right) \quad (12)$$

## 6 Bias of the Objective

We detail the bias induced by taking a subset of the samples in a group as mentioned in Section 3 of the main paper. We build an estimate of $\mathcal{L}(\mathcal{G}, \theta, \phi_c, \phi_s)$ using mini-batches of grouped observations.

$$\mathcal{L}(\mathcal{G}_b, \theta, \phi_c, \phi_s) = \frac{1}{|\mathcal{G}_b|} \sum_{G \in \mathcal{G}_b} \text{ELBO}(G; \theta, \phi_s, \phi_c) \quad (13)$$

If we take all observations in each group $G \in \mathcal{G}_b$, it is an unbiased estimate. However when the groups sizes are too large and we subsample $G$, this estimate is biased. The $\text{ELBO}(G; \theta, \phi_s, \phi_c)$ for a group is

$$\text{ELBO}(G; \theta, \phi_s, \phi_c) = \sum_{i \in G} \mathbb{E}_{q(C_G | \mathbf{X}_G; \phi_c)}[\mathbb{E}_{q(S_i | X_i; \phi_s)}[\log p(X_i | C_G, S_i; \theta)]] \\ - \sum_{i \in G} \text{KL}(q(S_i | X_i; \phi_s) || p(S_i)) - \text{KL}(q(C_G | \mathbf{X}_G; \phi_c) || p(C_G)). \quad (14)$$

Let us take a subsample $H$ of $G$ and consider the estimate $\text{ELBO}(G; \theta, \phi_s, \phi_c)_H$. The superscript $H$ denotes the fact that we use a subsample of $G$ to estimate $\text{ELBO}(G; \theta, \phi_s, \phi_c)_H$

$$\text{ELBO}(G; \theta, \phi_s, \phi_c)_H = \sum_{i \in H \subseteq G} \mathbb{E}_{q(C_G | \mathbf{X}_H; \phi_c)}[\mathbb{E}_{q(S_i | X_i; \phi_s)}[\log p(X_i | C_G, S_i; \theta)]] \\ - \sum_{i \in H \subseteq G} \text{KL}(q(S_i | X_i; \phi_s) || p(S_i)) - \text{KL}(q(C_G | \mathbf{X}_H; \phi_c) || p(C_G)). \quad (15)$$

where $q(C_G | \mathbf{X}_H; \phi_c)$ is computed using $\mathbf{X}_H$.

The gradient with respect to the parameters $\theta, \phi_s, \phi_c$ is

$$\nabla_\theta \text{ELBO}(G; \theta, \phi_s, \phi_c)_H = \sum_{i \in H \subseteq G} \mathbb{E}_{q(C_G | \mathbf{X}_H; \phi_c)}[\mathbb{E}_{q(S_i | X_i; \phi_s)}[\nabla_\theta \log p(X_i | C_G, S_i; \theta)]],$$

$$\nabla_{\phi_s} \text{ELBO}(G; \theta, \phi_s, \phi_c)_H = \sum_{i \in H \subseteq G} \mathbb{E}_{q(C_G | \mathbf{X}_H; \phi_c)}[\nabla_{\phi_s} \mathbb{E}_{q(S_i | X_i; \phi_s)}[\log p(X_i | C_G, S_i; \theta)]],$$

$$- \sum_{i \in H \subseteq G} \nabla_{\phi_s} \text{KL}(q(S_i | X_i; \phi_s) || p(S_i)) \quad (16)$$

$$\nabla_{\phi_s} \text{ELBO}(G; \theta, \phi_s, \phi_c)_H = \sum_{i \in H \subseteq G} \nabla_{\phi_c} \mathbb{E}_{q(C_G | \mathbf{X}_H; \phi_c)}[\mathbb{E}_{q(S_i | X_i; \phi_s)}[\log p(X_i | C_G, S_i; \theta)]] \\ - \nabla_{\phi_c} \text{KL}(q(C_G | \mathbf{X}_H; \phi_c) || p(C_G)).$$

Since we build the content variational approximation in a non-linear manner with the Product of Normals or the Mixture of Normals methods, the gradients $\nabla_\theta \text{ELBO}(G; \theta, \phi_s, \phi_c)_H$ and $\nabla_{\phi_c} \text{ELBO}(G; \theta, \phi_s, \phi_c)_H$ do not decompose in an unbiased manner, i.e. by summing the gradient of subsampled groups we do not retrieve the gradient computed using the entire group. The resulting bias will depend on the method employed. For future work, we plan on analysing the effect of the bias. In detail we want to derive a manner to correct the bias and verify that the biased estimator do not over-estimate the true objective function, that is the sum of the groups Evidence Lower Bound.

# 7 Stochastic Variational Inference (SVI) Results

We show in Figure 5 the qualitative results of Stochastic Variational Inference (SVI) Hoffman et al. [2013] on the swapping evaluation. We see that while SVI disentangles the style and the content, but the resulting image quality is poor. In the case of MS-Celeb-1M, we trained SVI for twice the number of epochs of the other models, that is in total 500 epochs, the training objective to maximise, that is the average group Evidence Lower Bound, remains lower than the ML-VAE model at the end of the training. This is because SVI does not share the parameters $\phi_c, \phi_s$ among observations at training, hence take a longer time to train. It is the first disadvantage of SVI compared to VAE-based model. At test-time, the SVI model requires expensive iterative inference. In the case of MS-Celeb-1M we used 200 epochs of test inference, it is possible that more epochs of test-time inference would have lead to better quality images but this already shows the limits of non-amortised variational inference and the advantages of the ML-VAE. We see that while SVI disentangles the style and the content, but the resulting image quality is poor.

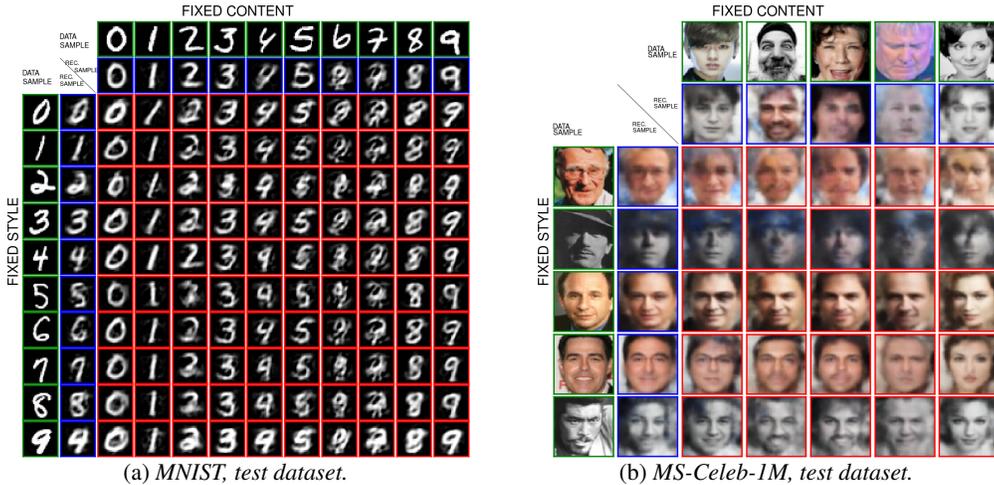

(a) *MNIST, test dataset.*   (b) *MS-Celeb-1M, test dataset.*

Figure 5: *Stochastic Variational Inference (SVI) Hoffman et al. [2013] qualitative results. Swapping, first row and first column are test data samples (green boxes), second row and column are reconstructed samples (blue boxes) and the rest are swapped reconstructed samples (red boxes). Each row is fixed style and each column is a fixed content. Best viewed in color on screen.*

# 8 Other Formulations Explored

We explored other possible formulations and we detail them here for the interested reader, along with the reasons for which we did not pursue them.

**SVI-Encode.** We explained the disavantadges of Stochastic Variational Inference (SVI), see Hoffman et al. [2013]. In order to remove the need for costly inference at test-time, we tried training an encoder to model the variational approximation of the latent representation $C, S$ corresponding to the generative model of the trained SVI model. We do not use training data but generated observations, sampling the latent representation from the prior. The encoder does not have any group information, and each sample has a separate content and style latent representation $C_i, S_i$. The model maximises the log-likelihood of the generative model under the latent code representation.

$$\sum_{i \in N} \mathbb{E}_{q(C_i, S_i | X_i; \phi)}[\log p(X_i | C_i, S_i, \theta)] \tag{17}$$

where $p(X_i|C_i, S_i, \theta)$ is the distribution corresponding to the generative model and $q(C_i, S_i|X_i; \phi)$ is the variational approximation. We refer to this method as SVI-Encode. The qualitative quality of this model on test samples is highly dependent on the quality of the generative model. Therefore it gives satisfactory qualitative results on MNIST, but poor qualitative results on MS-Celeb-1M where the qualitative results of SVI are not satisfactory. However, we think that a model trained alternatively between SVI and this SVI-Encode could benefit from the disentanglement power of SVI and amortised inference at test-time. We leave this for possible future work.

**Regularising the objective.** A possible formulation of the problem is to employ a regular VAE with an additional term to enforce observations within a group to have similar variational approximations of the content. This model separates the latent representation of the style, $S$ and the latent representation of the content $C$ but each observation $X_i$ within a group has its own latent variables $S_i$ and $C_i$. The sharing of the content within a group is enforced by adding a penalisation term based on a symmetrised Kullback-Leibler divergence between the content latent representation of the observations belonging to the same group. The resulting model maximises the objective[3]

$$
\begin{aligned}
& \frac{1}{|\mathcal{N}|} \sum_{i \in N} \mathbb{E}_{q(C_i, S_i | X_i; \phi)}[\log p(X_i | C_i, S_i, \theta)] \\
& - \frac{1}{|\mathcal{N}|} \sum_{i=1}^{N} \mathrm{KL}(q(S_i | X_i; \phi_s) \| p(S_i)) - \frac{1}{|\mathcal{N}|} \sum_{i=1}^{N} \mathrm{KL}(q(C_i | X_i; \phi_c) \| p(C_i)) \\
& - \frac{\lambda}{|\mathcal{G}|} \sum_{G \in \mathcal{G}} \frac{1}{|G|/2} \sum_{\substack{i \in [1, |G|/2], \\ s.t. X_{2i} \in G, X_{2i+1} \in G}} \frac{1}{2} \Big[ \mathrm{KL}(q(C_{2i} | X_{2i}, \phi_c) \| q(C_{2i+1} | X_{2i+1}, \phi_c)) \\
& + \mathrm{KL}(q(C_{2i+1} | X_{2i+1}, \phi_c) \| q(C_{2i} | X_{2i}, \phi_c)) \Big],
\end{aligned}
\tag{18}
$$

where $\lambda$ is an hyper-parameter to cross-validate. In our experiment, the drawback of this model is that if the latent representation size is too large (in detail, with the size $100$ used in our experiments for MS-Celeb-1M), the model sets the content latent representation to the prior $p(C)$ is order to encounter a null penalty. The observations are encoded in the style latent representation only. This is a known problem of VAE, see Chen et al. [2017]. On the opposite, the ML-VAE model is more robust to this problem.

---

[3]The equation was corrected compared to the submitted version.

# References

Xi Chen, Diederik P. Kingma, Tim Salimans, Yan Duan, Prafulla Dhariwal, John Schulman, Ilya Sutskever, and Pieter Abbeel. Variational lossy autoencoder. *ICLR*, 2017.

Yandong Guo, Lei Zhang, Yuxiao Hu, Xiaodong He, and Jianfeng Gao. MS-Celeb-1M: A dataset and benchmark for large scale face recognition. *ECCV*, 2016.

Matthew D. Hoffman, David M. Blei, Chong Wang, and John Paisley. Stochastic variational inference. *JMLR*, 2013.

Diederik P. Kingma and Jimmy Ba. Adam: A method for stochastic optimization. *ICLR*, 2015.

Yann Lecun, Léon Bottou, Yoshua Bengio, and Patrick Haffner. Gradient-based learning applied to document recognition. *Proceedings of the IEEE*, pages 2278–2324, 1998.


\end{document}